\documentclass{article}

\usepackage{microtype}
\usepackage{graphicx}
\usepackage{subfigure}
\usepackage{booktabs} 

\usepackage{hyperref}

\usepackage{algorithm} 
\usepackage{algorithmic}  
\usepackage[algo2e]{algorithm2e} 
\usepackage[dvipsnames]{xcolor}

\usepackage[accepted]{icml2024}

\usepackage{amsmath}
\usepackage{amssymb}
\usepackage{mathtools}
\usepackage{amsthm}

\usepackage[capitalize,noabbrev]{cleveref}

\newcommand\XS{\boldsymbol{X}_S}

\newcommand\X{\boldsymbol{X}}
\newcommand\xs{\boldsymbol{x}_S}

\newcommand\xsb{\boldsymbol{x}_{\bar{S}}}
\newcommand\x{\boldsymbol{x}}

\newcommand\A{\boldsymbol{a}}
\newcommand\YSt{\mathscr{Y}^{\star}}
\newcommand\zs{\boldsymbol{z}_S}

\newcommand{\niton}{\not\owns}

\usepackage{parskip}
\usepackage{dsfont}
\usepackage{mathrsfs}  

\usepackage{graphicx,array,tikz,multirow}
\usepackage{caption,subcaption} 
\usepackage{svg,float}
\usepackage{booktabs,paralist}
\newcolumntype{x}[1]{>{\centering\arraybackslash\hspace{0pt}}p{#1}}
\usepackage[section]{placeins}
\usepackage{hanging}

\theoremstyle{plain}
\newtheorem{theorem}{Theorem}[section]

\theoremstyle{definition}
\newtheorem{definition}[theorem]{Definition}

\theoremstyle{remark}

\usepackage[textsize=tiny]{todonotes}

\icmltitlerunning{Submission and Formatting Instructions for ICML 2024}

\begin{document}

\twocolumn[
\icmltitle{Local and Regional Counterfactual Rules: Summarized and Robust Recourses}



\icmlsetsymbol{equal}{*}

\begin{icmlauthorlist}
\icmlauthor{Salim I. Amoukou}{yyy,sch}
\icmlauthor{Nicolas J.B. Brunel}{yyy,comp}
\end{icmlauthorlist}

\icmlaffiliation{yyy}{University Paris Saclay, LaMME}
\icmlaffiliation{comp}{Quantmetry, ENSIIE}
\icmlaffiliation{sch}{Stellantis}

\icmlcorrespondingauthor{Salim I. Amoukou}{salimamoukou@gmail.com}

\icmlkeywords{Machine Learning, ICML}

\vskip 0.3in
]



\printAffiliationsAndNotice{\icmlEqualContribution} 

\begin{abstract}
Counterfactual Explanations (CE) face several unresolved challenges, such as ensuring stability, synthesizing multiple CEs, and providing plausibility and sparsity guarantees. From a more practical point of view, recent studies \citep{himanoisycounterfactuals} show that the prescribed counterfactual recourses are often not implemented exactly by individuals and demonstrate that most state-of-the-art CE algorithms are very likely to fail in this noisy environment. To address these issues, we propose a probabilistic framework that gives a sparse local counterfactual rule for each observation, providing rules that give a range of values capable of changing decisions with high probability. These rules serve as a summary of diverse counterfactual explanations and yield robust recourses. We further aggregate these local rules into a regional counterfactual rule, identifying shared recourses for subgroups of the data. Our local and regional rules are derived from the Random Forest algorithm, which offers statistical guarantees and fidelity to data distribution by selecting recourses in high-density regions. Moreover, our rules are sparse as we first select the smallest set of variables having a high probability of changing the decision. We have conducted experiments to validate the effectiveness of our counterfactual rules in comparison to standard CE and recent similar attempts. Our methods are available as a Python package.
\end{abstract}

\section{Introduction}
In recent years, many explanation methods have been developed for explaining machine learning models, with a strong focus on local analysis, i.e., generating explanations for individual prediction \citep{molnar2022}. Among these, Counterfactual Explanations \citep{Wachter2017CounterfactualEW} have emerged as a popular technique. In contrast to local attribution methods~ \citep{lundberg2020local2global, ribeiro2016why}, which assign importance scores to each feature, Counterfactuals Explanations (CE) describe the smallest modification to the feature values that changes the prediction to a desired target. These modifications are often called recourses. While CE can be intuitive and user-friendly, they have practical limitations. Most CE methods depend on gradient-based algorithms or heuristic approaches \citep{survey_counterfactual}, which may fail to identify the most natural modification and lack guarantees. Most algorithms either do not ensure sparsity (changes to the smallest number of features) or fail to generate plausible samples \citep{counterfactual_r1, CHOU202259}. Several studies \citep{optimalce_vidal, face_counterfactual, prototype_basedce} attempt to address the plausibility and the sparsity issues by incorporating ad-hoc constraints.

\begin{figure*}[ht!]
    \centering
    \includegraphics[scale=0.6]{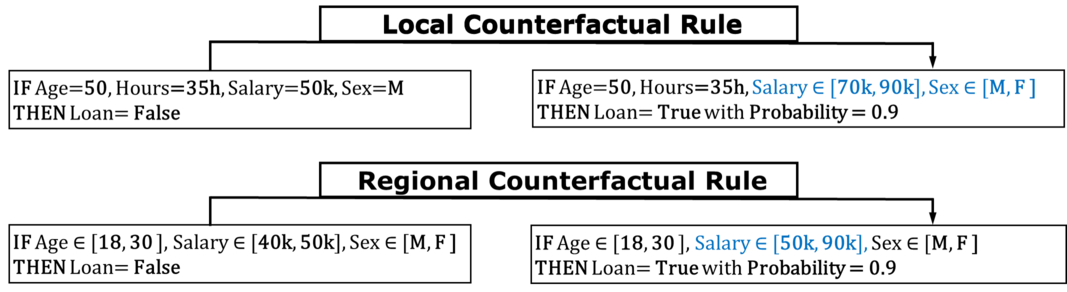}
    \caption{Illustration of local and regional Counterfactual Rules for a fictitious dataset with four variables: Age, Salary, Sex, and HoursPerWeek. Local rules change a single instance's decision, while regional rules apply to a sub-population. Blue indicates the suggested rules for changing decisions.}
    \label{fig:oce}
\end{figure*}

In another direction, numerous papers \citep{dice, Karimi2020ModelAgnosticCE, diverce_ce} encourage the generation of diverse counterfactuals to find actionable recourse \citep{Ustun2019ActionableRI}. Actionability is a vital desideratum, as some features may be non-actionable, and generating many counterfactuals increases the chance of getting actionable recourse. However, the diversity of CE compromises the intelligibility of the explanation, and the synthesis of various CE, in general, remains an unsolved challenge \citep{rethinkinxai}. Recently, \cite{himanoisycounterfactuals} highlights a new problem of CE called: \textit{noisy responses to prescribed recourses}. In real-world scenarios, some individuals may not be able to implement exactly the prescribed recourses, and they show that most CE methods fail in this setting. 

Consequently, we propose to reverse the usual way of explaining with counterfactual by computing \textit{Counterfactual rules}. We introduce a new line of counterfactuals, constructing interpretable policies (or rules) for changing a decision with a high probability while ensuring the stability of the derived recourse. These policies are sparse, faithful to the data distribution and their computation comes with statistical guarantees. Our proposal is to find a general rule that permits changing the decision while fixing some features instead of generating many counterfactual samples. These rules can be seen as a summary of the possible diverse counterfactual samples.  Additionally, we show that this method can be extended to create a common counterfactual policy for subgroups of the data, which aids model debugging and bias detection.  We introduce an algorithm to sample CEs or recourses from the generated rules. Notably, our approach is model-agnostic, meaning it does not need the model to make predictions or calculate other quantities, such as gradients. Instead, it is an inferential approach and relies solely on historical data. As a result, our approach can be applied not only to generate counterfactuals for a specific model but also in scenarios where we don't have access to the model or when making predictions from it is costly. Our method handles categorical and continuous features and classification and regression problems. In addition, we can use it to generate recourses directly for the data-generating process. An illustration of the counterfactual rules we introduce is illustrated in Figure \ref{fig:oce}.

\textbf{Notes on technical novelties.} While the Local Counterfactual Rule is a novel concept, the Regional Counterfactual Rule shares similarities with some recent works. Indeed, \cite{rawal2020beyond} proposed Actionable Recourse Summaries (AReS), a framework that constructs global counterfactual recourses to have a global insight into the model and detect unfair behavior. Despite similarities with the Regional Counterfactual Rule, there are notable differences. Our methods can handle regression problems and work directly with continuous features. AReS requires discretizing continuous features, leading to a trade-off between speed and performance~\citep{globalce}. Too few bins yield unrealistic recourse, while too many bins result in excessive computation time.  AReS employs a greedy heuristic search approach to find global recourse, which may result in unstable and inaccurate recourse. Our approaches overcome these limitations by leveraging on the informative partitions obtained from a Random Forest (RF), removing the need for an extensive search space, and focusing on high-density regions for plausibility. Additionally, we prioritize changes to the smallest number of features by identifying the smallest subset $S$ of variables $\XS$ and associated value ranges for each observation or subgroup that have the highest probability of changing the prediction. We compute this probability with a consistent estimator of the conditional distribution $Y | \boldsymbol{X}_S$ obtained from a RF.

\textbf{Our contributions can be summarised as follows:}
\begin{itemize}
    \item We redefine the problem of CE generation by introducing Counterfactual Rules that provide summarized and robust CE. We also introduce an algorithm to sample recourses efficiently from these rules.
    \item Our approach leverages the learned partitions of a Random Forest (RF) as an efficient tool for navigating high-dimensional spaces. It focuses on high-density regions, ensuring the generation of plausible CE. Additionally, it facilitates the identification of the minimal set of features requiring modification to change prediction towards desired outcomes with high probability.
    \item Our methods offer both local and regional explanations and handle regression and classification tasks.
    \item We quantitatively demonstrate the effectiveness of our approach across multiple desiderata, such as accuracy, plausibility, sparsity and robustness.
\end{itemize}

\section{Minimal Counterfactual Rules}
Consider a dataset $ \mathcal{D}_n = \{(\boldsymbol{X}_i,Y_i)\}_{i=1}^{n}$  consisting of i.i.d observations  of $ (\X, Y) \sim P_{\X}P_{Y|\X}$, where $ \X \in \mathcal{X}$ (typically $ \mathcal{X}\subseteq\mathbb{R}^p$) and $ Y \in \mathcal{Y}$. The output set $\mathcal{Y}$ can be either discrete or continuous. We denote $ [p] = \{1, \dots, p\}$, and for a given subset $ S \subseteq [p]$, $\XS = (X_i)_{i \in S}$ represents a subgroup of features, and we write $ \x=(\xs,\xsb)$.

For a given observation $(\x,y)$,  we consider a target set $\YSt \subset \mathcal{Y}$, such that  $y\notin \YSt$.  In the case of a classification problem, $\YSt = \{ y^\star\}$ is a singleton where $y^\star\in \mathcal{Y}$ and $y^\star \neq y$. Unlike conventional approaches, our definition of CE also accommodates regression problems by considering $\YSt = [a,b]\subset \mathbb{R}$, and the definitions and computations remain the same for both classification and regression. The classic CE problem, defined here only for classification, considers a predictor $f:\mathcal{X} \rightarrow \mathcal{Y}$, trained on dataset $\mathcal{D}_n$ and search a function $\A: \mathcal{X} \rightarrow \mathcal{X}$, such that for all observations $\x \in \mathcal{X}$, $f(\x)\neq y^\star$, we have $f(\A(\x))=y^\star$. The function is defined  point-wise by solving an optimisation program. Most often $\A(\cdot)$ is not a single-output function, as $\A(x)$ may be in fact a collection of (random) values $\{\x_1^{CF},\dots,\x_k^{CF}\}$, which represent the counterfactual samples. A more recent perspective, proposed by \cite{cet4}, defines $\A$ as a decision tree, where for each leaf $L$, a common action is predicted for all instances $\x \in L$ to change their predictions.

Our approach diverges slightly from the traditional model-based definition of CE as we can directly consider observation $(\X, Y)$ rather than model prediction $(\X, f(\X))$. To illustrate the concept, let's consider a binary classification problem, where the input space can be divided into two regions $R_0, R_1$. These correspond to the support of the distributions $\X | Y=0$ and $\X| Y=1$. These regions may not be disjoint or convex spaces and can be represented as a union of several sets. Given an observation $\x = (\x_S, \x_{\bar{S}})$ with label $y=0$, our method consists of finding the minimal subset of variables $S \subseteq \{1, \dots, p\}$ to move $\x$ by modifying $\xs$ into a set within the region $R_1$. The objective is to move $\x$ to a high-density set with low variance with respect to the target variable $Y$, while altering as few variables as possible. 

\begin{figure}[ht!]
    \centering
    \includegraphics[scale=0.2]{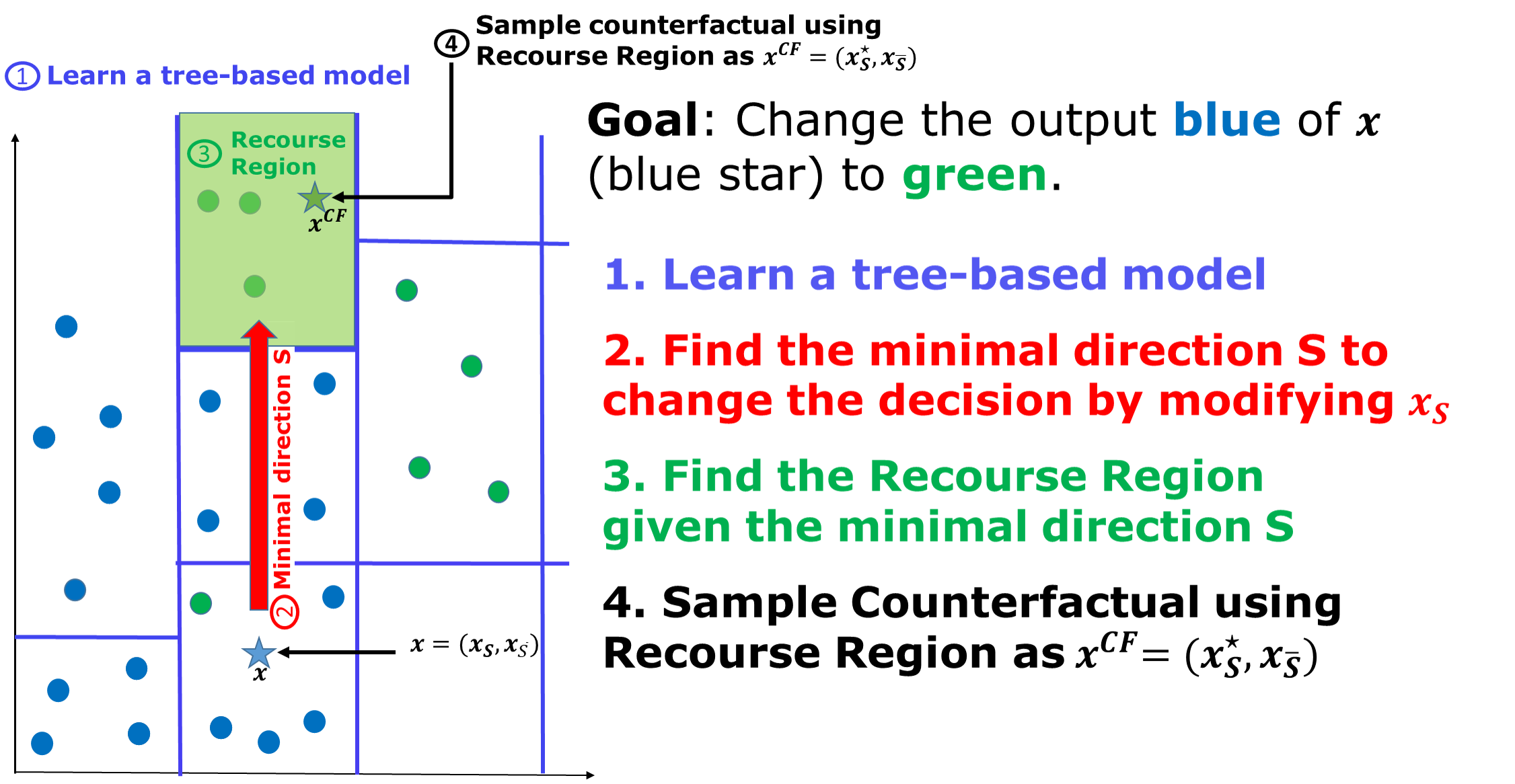}
    \caption{Illustration of the 4-stages in our methodology for computing sparse counterfactuals}
    \label{fig:cr_demonstration}
\end{figure}

Figure \ref{fig:cr_demonstration} provides a visual representation of our approach in the binary case. The first step involves learning a tree-based model on our data, enabling us to partition the input space based on the target variable $Y$. By examining the tree leaves, we can easily identify the optimal direction $S$ to modify the decision and the target region corresponding to the counterfactual rule. Moreover, these leaves can serve not only as rules but also as a means to generate recourses.

Our approach is different from the optimization approach for generating recourse, as it is model-agnostic, meaning it does not require the model to generate further predictions or compute gradients. This flexibility allows us to apply our approach to generate recourse either for the predictions of a given model $(\X, f(\X))$ or the data-generating process $(\X, Y)$. A model-agnostic approach was also proposed by \citep{black2020fliptest, de2021transport} under the name of transport-based counterfactuals. It consists of finding a map $T$ between the distribution of $\X|Y=0$ and $\X|Y=1$ such that each observation of class $Y=0$ is linked to the most similar observation of class $Y=1$. \citet{de2021transport} shows that it coincides with causal counterfactual under appropriate assumptions. In the following discussion, we consider the data $(\X, Y)$ for the presentation of the methods, although they can also be applied to generate recourses for a model prediction $(\X, f(\X))$ as well.

Our approach is hybrid, as we do not suggest a single action for each observation or subspace of $\mathcal{X}$ but provide sets of possible perturbations. A Local Counterfactual Rule (L-CR) for target $\YSt$ and observation $(\x, y)$ (with $y\notin \YSt$) is a rectangle $C_{S}(\boldsymbol{x}, \YSt) = \prod_{i\in S} [a_i, b_i], a_i, b_i \in \overline{\mathbb{R}}$ such that for all counterfactual samples of $\x=\left(\xs,\xsb \right)$ obtained as $\boldsymbol{x}^{CF} = \left(\zs,\xsb \right)$ with $\zs \in C_{S}(\x,\YSt)$ and $\boldsymbol{x}^{CF}$ an in-distribution sample, then $y^{CF}$ is in  $\YSt$ with a high probability, where $y^{CF}$ is the output of $\x^{CF}$ given by the model $f$ or the data-generating process. Similarly, a Regional Counterfactual Rule (R-CR) $C_S(\boldsymbol{R}, \YSt)$ is defined for target $\YSt$ and a rectangle $\boldsymbol{R}=\prod_{i=1}^{d} [a_i, b_i], a_i, b_i \in \overline{\mathbb{R}}$, which represent a subspace of $\mathcal{X}$ of similar observations, if for all observations $\x=(\xs,\xsb) \in \boldsymbol{R}$, the countefactual samples obtained as $\boldsymbol{x}^{CF} = (\zs,\xsb)$ with $\zs \in C_S(\boldsymbol{R},\YSt)$ and $\boldsymbol{x}^{CF}$ an in-distribution sample are such that $y^{CF}$ is in $\YSt$ with high probability. Our approach constructs such rectangles in a sequential manner. Firstly, we identify the minimal directions $S \subseteq [p]$ that offer the highest probability of changing the decision. Next, we determine the optimal intervals $[a_i,b_i]$ for $i \in S$ that change the decision to the desired target. Additionally, we propose a method to derive traditional Counterfactual Explanations (CE) (i.e., actions that alter the decision) or recourses using our Counterfactual Rules. A central tool in this approach is the Counterfactual Decision Probability presented below.
\begin{definition}
\label{def:cdp}\textbf{Counterfactual Decision Probability (CDP).} The Counterfactual Decision Probability of the subset $S\subseteq \{1, \dots, p\} $,
w.r.t $\boldsymbol{x}=\left(\boldsymbol{x}_{S},\boldsymbol{x}_{\bar{S}}\right)$, output $y$ and the desired target $\YSt$ (s.t. $y \notin \YSt)$ is 
\begin{align}
    CDP_{S}\left(\x, \YSt\right)=\mathbb{P}\left(Y \in  \YSt\left|\boldsymbol{X}_{\bar{S}}=\boldsymbol{x}_{\bar{S}}\right.\right).    
\end{align}
\end{definition} 
The $CDP$ of the subset S is the probability that the decision changes to the desired target $\YSt$ by sampling the features $\XS$ given $\boldsymbol{X}_{\bar{S}} = \boldsymbol{x}_{\bar{S}}$. It is related to the Same Decision Probability  $SDP_{S}(\mathscr{Y}; \boldsymbol{x}) = \mathbb{P}\left(Y \in \mathscr{Y} \vert \XS=\xs \right)$ used in \citep{amoukou2021consistent} for solving the dual problem of selecting the most local important variables for obtaining and maintaining the decision  $Y \in \mathscr{Y}$, where $\mathscr{Y}\subset \mathcal{Y}$. The set $S$ is called the Minimal Sufficient Explanation. Indeed, we have $CDP_S(\x, \YSt) = SDP_{\overline{S}}(\x, \YSt)$. The computation of these probabilities is challenging and discussed in Section \ref{sec:estimation}. Next, we define the minimal subset of features $S$ that allows changing the decision to the target set with a given high probability $\pi$.%

\begin{definition}  \label{def:minimal_countset}(\textbf{Minimal Divergent Explanations}). Given an instance $(\boldsymbol{x}, y)$ and a desired target $\YSt \niton y$, $S$ is a Divergent Explanation for probability $\pi>0$ if $CDP_{S}\left(\x, \YSt\right)\geq\pi$, and no subset $Z$ of $S$ satisfies $CDP_{Z}\left(\x, \YSt\right)\geq\pi$. Hence, a Minimal Divergent Explanation is a Divergent Explanation with the smallest size.


\end{definition}
The set satisfying these properties is not unique, and we can have several Minimal Divergent Explanations. Note that the probability $\pi$ represents the minimum level required for a set to be chosen for generating counterfactuals, and its value should be as high as possible and depends on the use case. With these concepts established, we can now define our main criterion for constructing a Local Counterfactual Rule (L-CR).
\begin{definition}\label{def:local_counterfactual_rule}  (\textbf{Local Counterfactual Rule}). Given an instance $(\boldsymbol{x}, y)$, a desired target $ \YSt \niton y$ , a Minimal Divergent Explanation $S$, the rectangle $\small C_{S}(\boldsymbol{x}, \YSt) = \prod_{i\in S} [a_i, b_i], a_i, b_i \in \overline{\mathbb{R}}$ is a Local Counterfactual Rule with probability $\pi_C$ if  $\small C_S(\x, \mathscr{Y}^\star) = \arg \max _{C} \mathbb{P}_{P_{\X}}\big(\boldsymbol{X}_S \in C \;\vert\;  \boldsymbol{X}_{\bar{S}} = \boldsymbol{x}_{\bar{S}}\big)$ such that $\small  CRP_S\Big(\x, \YSt\Big) = \mathbb{P}( Y \in \YSt \; | \;\boldsymbol{X}_S \in C_S(\boldsymbol{x}, \YSt),\; \boldsymbol{X}_{\bar{S}} = \x_{\bar{S}})$ satisfies
    \begin{equation}
        CRP_S\Big(\x, \YSt\Big)  \geq \pi_C.
    \end{equation}

$ \mathbb{P}_{P_{\X}}\big(\boldsymbol{X}_S \in C_S(\x, \YSt)\;\vert \;\boldsymbol{X}_{\bar{S}} = \boldsymbol{x}_{\bar{S}}\big)$ represent the plausibility of the rule and by maximizing it, we ensure that the rule lies in a high-density region. $CRP_S$ is the Counterfactual Rule Probability. The higher the probability $\pi_C$ is, the better the relevance of the rule $C_S(\x, \YSt)$ is for changing the decision to the desired target. 
\end{definition}

In practice, we often observe that the Local CR $C_{S}(\cdot, \YSt)$ for neighboring observations $\x$ and $\x^{\prime}$ are quite similar, as the Minimal Divergent Explanations tend to be alike, and the corresponding hyperrectangles frequently overlap. This observation motivates a generalization of these Local CRs to hyperrectangles $\boldsymbol{R} = \prod_{i=1}^{d} [a_i, b_i], a_i, b_i \in \overline{\mathbb{R}}$, which group together similar observations. We denote $\text{supp}(\boldsymbol{R}) = \{i : [a_i, b_i] \neq \overline{\mathbb{R}}\}$ as the support of the rectangle and extend the Local CRs to Regional Counterfactual Rules (R-CR). To achieve this, we denote $\boldsymbol{R}_{\bar{S}} = \prod_{i \in \bar{S}} [a_i, b_i]$ as the rectangle with intervals of $\boldsymbol{R}$ in $\text{supp}(\boldsymbol{R}) \cap \bar{S}$, and define the corresponding Counterfactual Decision Probability (CDP) for rule $\boldsymbol{R}$ and subset $S$ as $CDP_S(\boldsymbol{R}, \YSt) = \mathbb{P}\left(Y \in \YSt \left|\boldsymbol{X}_{\bar{S}} \in \boldsymbol{R}_{\bar{S}}\right.\right)$. Consequently, we can compute the Minimal Divergent Explanation for rule $\boldsymbol{R}$ using the corresponding CDP for rules, following Definition (\ref{def:minimal_countset}). The Regional Counterfactual Rules (R-CR) correspond to Definition (\ref{def:local_counterfactual_rule}) with the associated CDP for rules.

\section{Estimation of the $CDP$ and $CRP$} \label{sec:estimation}
To compute the probabilities $CDP_S$ and $CRP_S$ for any $S$, we use a dedicated Random Forest (RF) that learns to predict the output of the model or the data-generating process. Indeed, the conditional probabilities $CDP_S$ and $CRP_S$ can be easily computed from a RF by combining the Projected Forest algorithm \citep{benard2021shaff} and the Quantile Regression Forest \citep{meinshausen2006quantile}. As a result, we can estimate the probabilities $CDP_S(\x, \mathscr{Y}^\star)$ consistently. This method has been previously utilized by \citep{amoukou2021consistent} for calculating the Same Decision Probability $SDP_S$.

\subsection{Projected Forest and $CDP_S$}
The estimator of the $SDP_S$ is based on the Random Forest \citep{breiman1984classification} algorithm. Assuming that we have trained a RF $m(\cdot)$ using the dataset $\mathcal{D}_n$, the model consists of a collection of $k$ randomized trees (for a detailed description of decision trees, see \citep{Loh2011ClassificationAR}). For each instance $\boldsymbol{x}$, the predicted value of the $l$-th tree is denoted as $m_l(\boldsymbol{x};\; \Theta_l)$, where $\Theta_l$ represents the resampling data mechanism in the $j$-th tree and the subsequent random splitting directions. The predictions of the individual trees are then averaged to produce the prediction of the forest as $ m(\boldsymbol{x}; \; \Theta_{1}, \dots, \Theta_{k}) = \frac{1}{k} \sum_{l=1}^{k} m_l(\boldsymbol{x};\; \Theta_l)$.
The RF can also be interpreted as an adaptive nearest neighbor predictor \citep{lin2006random, biau_layered_2010} or kernel methods \citep{breiman2000some, geurts2006extremely, scornet2016random}. For every instance $\boldsymbol{x}$, the observations in $\mathcal{D}_n$ are weighted by $w_{n, i}(\boldsymbol{x})$, with $i=1, \dots, n$. As a result, the prediction of the RF can be reformulated as $ m(\boldsymbol{x}; \; \Theta_{1}, \dots, \Theta_{k}) = \sum_{i=1}^{n} w_{n, i}(\boldsymbol{x}) Y_i.$
This emphasizes the central role played by the weights in the RF's algorithm. See \citep{meinshausen2006quantile, amoukou2021consistent} for a detailed description of the weights. Consequently, it naturally gives estimators for other quantities, e.g., cumulative hazard function \citep{ishwaran2008random}, treatment effect \citep{wager2017estimation, jocteur2023heterogeneous}, conditional density  \citep{du2021wasserstein}. For instance, \cite{meinshausen2006quantile} showed that we can use the same weights to estimate the conditional distribution function with the following estimator $ \widehat{F}(y | \boldsymbol{X} = \boldsymbol{x}) = \sum_{i=1}^{n} w_{n, i}(\boldsymbol{x}) \mathds{1}_{Y_{i} \leq y}$.
In another direction, \cite{benard2021shaff} introduced the Projected Forest algorithm \citep{benard2021mda, benard2021shaff} that aims to estimate $E[Y | \boldsymbol{X}_S]$ by modifying the RF's prediction algorithm.

\begin{figure*}[ht!]
\centering
\subfigure[]{\includegraphics[width=0.2\linewidth]{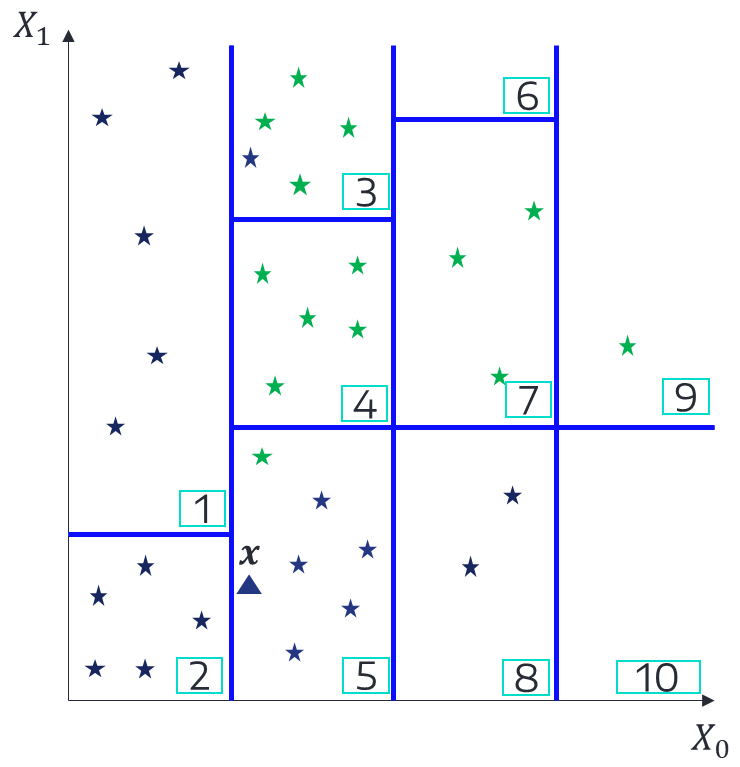}\label{fig:forest_part}}
\subfigure[]{\includegraphics[width=0.2\linewidth]{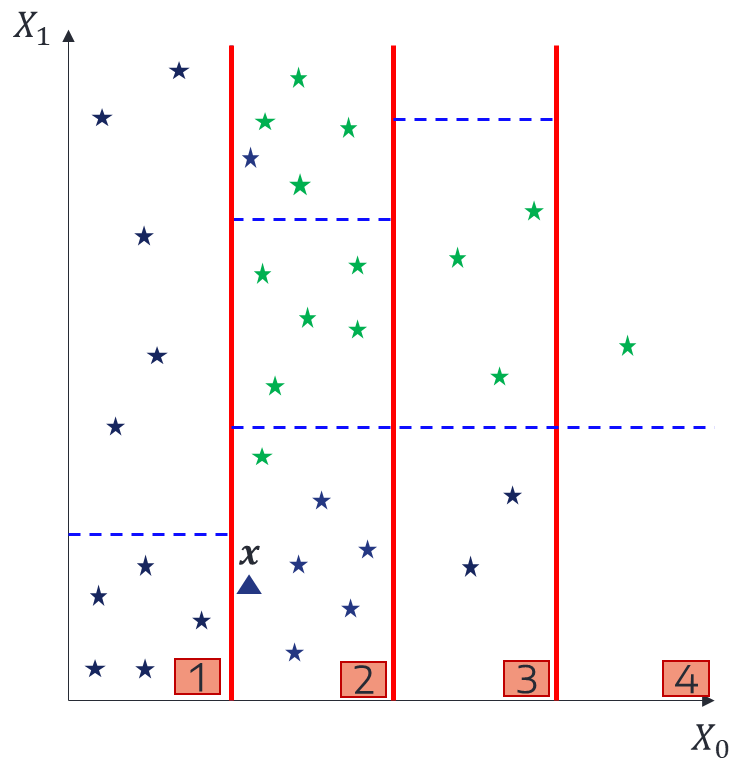}\label{fig:projected_part}}
\subfigure[]{\includegraphics[width=0.2\linewidth]{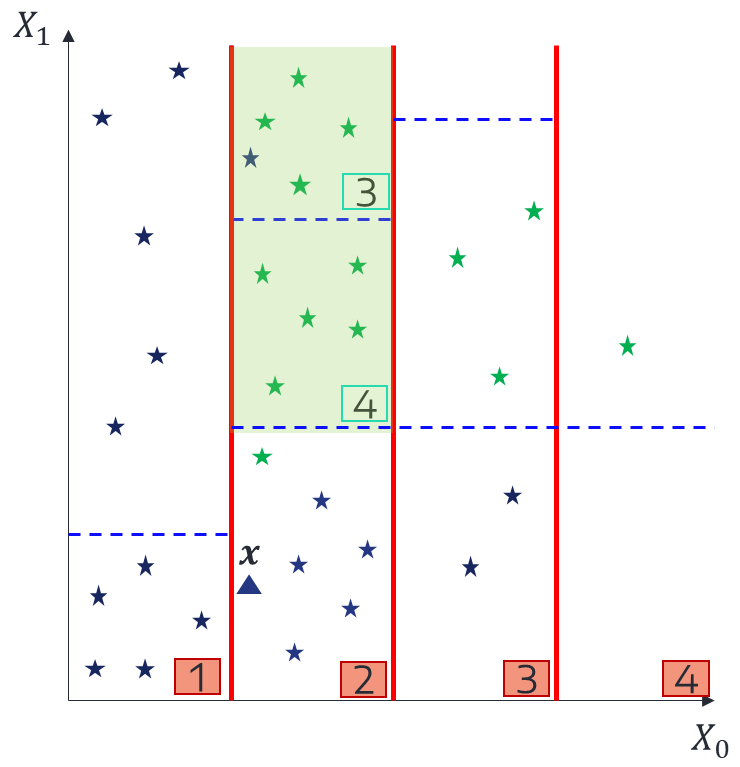}\label{fig:cr}}
\caption{(a) Partition of the Random Forest, (b) Partition of the Projected Random Forest when we condition given $X_0$, i.e., ignoring the splits on $X_1$, (c) The optimal Counterfactual Rule of $\boldsymbol{x}$ when we condition given $X_0=x_0$ is the green region.}
\end{figure*}

\textbf{Projected Forest:}  To estimate $E[Y | \XS = \xs]$ instead of $E[Y | \boldsymbol{X} = \x]$ using a RF, \cite{benard2021interpretable} suggests to simply ignore the splits based on the variables not contained in $S$ from the tree predictions. More formally, it consists of projecting the partition of each tree of the forest on the subspace spanned by the variables in S.  The authors also introduced an algorithmic trick that computes the output of the Projected Forest efficiently without modifying the initial tree structures. It consists of dropping the observations down in the initial trees, ignoring the splits that use a variable not in $S$: when it encounters a split involving a variable $i \notin S$, the observations are sent to the left and right children nodes. Therefore, each instance falls in multiple terminal leaves of the tree. To compute the prediction of $\xs$, we follow the same procedure and gather the set of terminal leaves where $\xs$ falls. Next, we collect the training observations which belong to every terminal leaf of this collection, in other words, we keep only the observations that fall in the intersection of the leaves where $\xs$ falls. Finally, we average their outputs $Y_i$ to estimate $E[Y | \XS = \xs]$. The authors show this algorithm converges asymptotically to the true projected conditional expectation $E[Y | \XS = \xs]$ under suitable assumptions. As the RF, the Projected Forest (PRF) assigns a weight to each training observation. The associated PRF is denoted $
m^{(S)}(\xs) = \sum_{i=1}^{n} w_{n, i}(\xs) Y_i$. Therefore, as the weights of the original forest was used to estimate the CDF, \cite{amoukou2021consistent} used the weights of the Projected Forest Algorithm to estimate $SDP$ as $\widehat{SDP}_{S}\left(\x, \mathscr{Y}^\star\right) = \sum_{i=1}^{n} w_{n, i}(\xs) \mathds{1}_{Y_i \in  \mathscr{Y}^\star}$. The idea is essentially to replace $Y_i$ by $\mathds{1}_{Y_i \in \mathscr{Y}^\star}$ in the Projected Forest equation defined above. \cite{amoukou2021consistent} also show that this estimator converges to the true $SDP_S$ under suitable assumptions and works very well in practice. Especially for tabular data, where tree-based models are known to perform well \citep{grinsztajn2022tree}. Similarly, we can estimate the $CDP$ with statistical guarantees \citep{amoukou2021consistent} using the following estimator $ \widehat{CDP}_{S}\left(\x, \YSt\right) = \sum_{i=1}^{n} w_{n, i}(\xsb) \mathds{1}_{Y_i \in \YSt}$.

\textbf{Remark:} We only give the estimator of $CDP_S$ of an instance $\x$. The estimator for $CDP_S$ of a rule $R$ will be discussed in the next section, as it is closely related to the estimator of the $CRP_S$.

\subsection{Regional RF and $CRP_S$}
Here,  we focus on estimating the $\small CRP_S(\x, \YSt) = \mathbb{P}(Y \in  \YSt \; | \boldsymbol{X}_S \in C_S(\boldsymbol{x}, \YSt), \; \boldsymbol{X}_{\bar{S}} = \boldsymbol{x}_{\bar{S}})$ and $\small CRP_S(\boldsymbol{R}, \YSt) = \mathbb{P}(Y \in \YSt \; | \boldsymbol{X}_S \in C_S(\boldsymbol{R};\YSt),\; \boldsymbol{X}_{\bar{S}} \in \boldsymbol{R}_{\bar{S}})$. For ease of reading, we remove the dependency of the rectangles $C_S$ in $\YSt$. Based on the previous section, we already know that the estimators using the RF will take the form of $ \widehat{CRP}_{S}\left(\x, \YSt\right) = \sum_{i=1}^{n} w_{n, i}^R(\x) \mathds{1}_{Y_i \in \YSt}$, so we only need to determine the appropriate weighting. The main challenge lies in the fact that we have a condition based on a region, e.g., $ \XS \in C_S(\boldsymbol{x})$ or $ \boldsymbol{X}_{\bar{S}} \in \boldsymbol{R}_{\bar{S}}$ (regional-based) instead of a condition of type $\XS = \xs$ (fixed value-based) as usual. However, we introduced a natural extension of the RF algorithm to handle predictions when the conditions are both regional-based and fixed value-based. As a result, cases with only regional-based conditions can be naturally derived.

\textbf{Regional RF to estimate $ CRP_S(\x, \YSt) = \mathbb{P}(Y \in \YSt \; | \; \boldsymbol{X}_S \in C_S(\boldsymbol{x}),\; \boldsymbol{X}_{\bar{S}} = \boldsymbol{x}_{\bar{S}})$.} The algorithm is based on a slight modification of RF and works as follows: we drop the observations in the trees, if a split used variable $i \in \bar{S}$, i.e., fixed value-based condition, we use the classic rules of RF, if $x_i \leq t$, the observations go to the left children, otherwise the right children. However, if a split used variable $i \in S$, i.e, regional-based condition, we use the rectangles $C_S(\boldsymbol{x}) = \prod_{i=1}^{|S|} [a_i, b_i]$. The observations are sent to the left children if $b_i \leq t$, right children if $a_i > t$ and if $t \in [a_i, b_i]$, the observations are sent both to the left and right children. Consequently, we use the weights of the Regional RF algorithm $w_{n, i}^R(\x)$ to estimate $CRP_S$, the estimator is $\widehat{CRP}_S(\x, \mathscr{Y}^\star) = \sum_{i=1}^{n} w_{n, i}^R(\x) \mathds{1}_{Y_i \in \mathscr{Y}^\star}$. In addition, the number of observations at the leaves is used as an estimate of $\mathbb{P}(\boldsymbol{X}_S \in C_S(\boldsymbol{x})\; \vert \; \boldsymbol{X}_{\bar{S}} = \boldsymbol{x}_{\bar{S}})$. A more comprehensive description and discussion of the algorithm are provided in the Appendix (\ref{sup:regional_forest}).

To estimate the $CDP$ of a rule $CDP_{S}\left(\boldsymbol{R}, \YSt \right)=\mathbb{P}\left(Y \in \YSt \left|\boldsymbol{X}_{\bar{S}}\in \boldsymbol{R}_{\bar{S}} \right.\right)$, we just have to apply the Projected Forest algorithm to the Regional RF, i.e., when a split involving a variable outside of $\bar{S}$ is met, the observations are sent both to the left and right children nodes, otherwise we use the Regional RF split rule, i.e., if an interval of $\boldsymbol{R}_{\bar{S}}$ is below $t$, the observations go to the left children, otherwise the right children and if $t$ is in the interval, the observations go to the left and right children. The estimator of the $CRP_S( \boldsymbol{R}, \YSt) = \mathbb{P}(Y \in \YSt \; | \boldsymbol{X}_S \in C_S(\boldsymbol{R};\YSt),\; \boldsymbol{X}_{\bar{S}} \in \boldsymbol{R}_{\bar{S}})$ for rule $\boldsymbol{R}$ is also derived from the Regional RF. Indeed, it is a special case of the Regional RF algorithm where there are only regional-based conditions.

\section{Learning the Counterfactual Rules}
The computation of the Local and Regional CR is performed using the estimators introduced in the previous section. First, we determine the Minimal Divergent Explanation, akin to the Minimal Sufficient Explanation \citep{amoukou2021consistent}, by exploring the subsets obtained using the $K = 10$ most frequently selected variables in the Random Forest estimator. $K$ is a hyper-parameter to choose according to the use case and computational power. We can also use any importance measure.  An alternative strategy to exhaustively searching through the $2^K$ possible subsets would be to sample a sufficient number of subsets, typically a few thousand, that are present in the decision paths of the trees in the forest. By construction,  these subsets are likely to contain influential variables. A similar strategy was used in \citep{basu2018iterative, benard2021shaff}. 

 Given an instance $\boldsymbol{x}$ or rectangle $\boldsymbol{R}$, target set $\YSt$ and their corresponding Minimal Divergent Explanation $S$, our objective is to find the maximal rule $C_S(\cdot) = \prod_{i \in S} [a_i, b_i]$ s.t. given $ \boldsymbol{X}_{\bar{S}} = \boldsymbol{x}_{\bar{S}}$ or $ \boldsymbol{X}_{\bar{S}} \in \boldsymbol{R}_{\bar{S}}$, and $ \XS \in C_S(\cdot)$, the probability that $ Y \in \YSt$ is high. Formally, we want: $ \mathbb{P}(Y \in \YSt | \XS \in C_S(\boldsymbol{x}), \boldsymbol{X}_{\bar{S}} = \boldsymbol{x}_{\bar{S}})$ or $ \mathbb{P}(Y \in \YSt| \XS \in C_S(\boldsymbol{R}), \boldsymbol{X}_{\bar{S}} \in \boldsymbol{R}_{\bar{S}} )$ above $ \pi_C$.
 
 The rectangles $C_S(\cdot) = \prod_{i\in S} [a_i, b_i]$ defining the CR are derived from the RF. In fact, these rectangles naturally arise from the partition learned by the RF. AReS \citep{rawal2020beyond}, on the other hand, relies on binned variables to generate candidate rules, testing all possible rules to select the optimal one. By leveraging the partition learned by the RF, we overcome both the computational burden and the challenge of choosing the number of bins. Moreover, by focusing only on the non-empty leaves containing training observations, we significantly reduce the search space. This approach allows identifying high-density regions of the input space to generate plausible counterfactual explanations.

 To illustrate the idea, we use a two-dimensional data $(X_0, X_1)$ with binary label Y represented as green and blue stars in Figure \ref{fig:forest_part}. We fit a Random Forest to classify this dataset and show its partition in Figure \ref{fig:forest_part}. We consider an instance $\boldsymbol{x}$ (blue triangle), and our goal is to change its classification from blue to green. From a visual analysis of cells/leaves of the RF, we deduce that the Minimal Divergent Explanation of $\boldsymbol{x}$ is $S = X_1$.  In Figure \ref{fig:projected_part}, we observe the leaves of the Projected Forest when not conditioning on $S = X_1$, thus projecting the RF's partition only on the subspace $X_0$. It consists of ignoring all the splits in the other directions (here the $X_1$-axis), thus $\boldsymbol{x}$ falls in the projected leaf 2 (see Figure \ref{fig:projected_part}) and its $CDP$ is  $CDP_{X_1}(\text{green}; \boldsymbol{x})=\frac{10 \text{ green}}{10\text{ green} + 17\text{ blue}} = 0.58$. To find the optimal rectangle $C_S(\boldsymbol{x}) = [a_i, b_i]$ in the direction of $X_1$, such that the decision changes, we can utilize the leaves of the RF. By looking at the leaves of the RF (Figure \ref{fig:forest_part})  for observations belonging to the projected RF leaf 2 (Figure \ref{fig:projected_part}) where $\boldsymbol{x}$ falls, we observe in Figure \ref{fig:cr} that the optimal rectangle for changing the decision, given $X_0 = x_0$ or being in the projected RF leaf 2, is the union of the intervals on $X_1$ of the leaf 3 and 4 of the RF (see the green region in Figure \ref{fig:cr}).
 
Given an instance $\boldsymbol{x}$ and its Minimal Divergent Explanation $S$, the first step is to collect observations that belong to the leaf of the Projected Forest given $\bar{S}$, where $\boldsymbol{x}$ falls. These observations correspond to those with positive weights in the computation of $ CDP_S(\x, \YSt) = \sum_{i=1}^{n} w_{n, i}^R(\boldsymbol{x}_{\bar{S}}) \mathds{1}_{Y_i \in \YSt}$, i.e., $ \{\X_i: w_{n, i}^R(\boldsymbol{x}_{\bar{S}}) >0\}$. Then we use the partition of the original forest to find the possible leaves in the direction $S$. The possible leaves are among the RF's leaves of the collected observations $ \{\boldsymbol{X}_i: w_{n,i}^R(\boldsymbol{x}_{\bar{s}}) >0\}$. Let denote $ L(\boldsymbol{X}_i)$ the leaf of the observation $\X_i$ with $ w_{n, i}(\boldsymbol{x}_{\bar{S}}) >0$. A possible leaf is a leaf $ L(\boldsymbol{X}_i)$ s.t. $ \mathbb{P}( Y \in \YSt | \XS \in L(\X_i)_S, \boldsymbol{X}_{\bar{S}} = \boldsymbol{x}_{\bar{S}}) \geq \pi_C$. Finally, we merge all the possible neighboring leaves to get the largest rectangle, and this maximal rectangle is the counterfactual rule. It is important to note that the union of possible leaves is not necessarily a connected space, which may result in multiple disconnected counterfactual rules.

We apply the same approach to find the regional CR. Given a rule $\boldsymbol{R}$ and its Minimal Divergent Explanation $S$, we used the Projection given $\boldsymbol{X}_{\bar{S}} \in \boldsymbol{R}_{\bar{S}}$ to identify compatible observations and their leaves. We then combine the possible ones that satisfy $CRP_S(\boldsymbol{R}, \YSt) \geq \pi_C$ to obtain the regional CR. For instance, if we consider Leaf 5 of the original forest as a rule (i.e., if $\boldsymbol{X} \in$ Leaf 5, then predict blue), its Minimal Divergent Explanation is also $S=X_1$. The Regional CR would be the green region in Figure \ref{fig:cr}. Indeed, satisfying the $X_0$ condition of Leaf 5 and the $X_1$ condition of Leaves 3 and 4 would cause the decision to change to green.

\section{Sampling CE using the CR}\label{sec:sampling_ce}

  Our approaches cannot be directly compared with traditional CE methods, as they return counterfactual samples, whereas we provide rules (ranges of vector values) that permit changing the decision with high probability. In some applications, users might prefer recourse to CR. Hence, we adapt the CR to generate counterfactual samples using a generative model. For example, given an instance $\x = (\xs, \x_{\bar{S}})$, target set $\YSt$ and its counterfactual rule $C_S(\x, \YSt)$, we want to find a sample $x^{CF} = (\boldsymbol{z}_S, \x_{\bar{S}})$ with $\boldsymbol{z}_S \in C_S(\x, \YSt)$ such that  $\x^{CF}$ is a realistic sample and $y^{CF} \in \YSt$. 
Instead of using a complex conditional generative model as \citep{modeling_td, sdv}, which can be difficult to calibrate, we use an energy-based generative approach \citep{ebmduvenaud, yanebm}. The core idea is to find $\boldsymbol{z}_S \in C_S(\x, \mathscr{Y}^\star)$ such that $\x^\star$ maximizes a given energy score, ensuring that $\x^\star$ lies in a high-density region. We use the negative outlier score of an Isolation Forest \citep{liu2008isolation} and Simulated Annealing \citep{review_simulated_annealing} to maximize the negative outlier score using the information of the counterfactual rules $C_S(\x, \YSt)$. In fact, the range values given by the CR $C_S(\x, \YSt)$ reduce the search space for $\boldsymbol{z}_S$ drastically. We used the marginal law of $\X$ given $\XS \in C_S(\x, \YSt)$ as the proposal distribution, i.e., we draw a candidate $\boldsymbol{z}_S$ by independently sampling each variable using the marginal law $\boldsymbol{z}_S \sim \prod_{i \in S} P_{X_j \; | \X_S \in C_S(\x, \YSt)}$ until we find an observation $\x^{CF} = (\boldsymbol{z}_S, \x_{\bar{S}})$ with high energy. The algorithm works similarly for sampling CE with the Regional CR. The methodology is described below in Appendix \ref{sec:algo}.

\section{Experiments}

\begin{table*}[ht!]
\caption{Results of the \textit{Accuracy} (Acc), \textit{Plausibility} (Psb), \textit{Sparsity} (Sprs), and \emph{Cost} (Cost) of the different methods. We compute each metric according to the positive (Pos) and negative (Neg) class. The blue value corresponds to the metric for the positive class (recourse for label=1 to label=0), and the red for the negative class (recourse for label=0 to label=1).}
\label{tab:results}
\tiny
\begin{tabular}{lllllllllllll}
                    & \multicolumn{4}{c}{\textbf{Compas}}                                                                            & \multicolumn{4}{c}{\textbf{Diabetes}}                                                                          & \multicolumn{4}{c}{\textbf{Nhanesi}}                                                                           \\ \cline{2-13} 
                    & \multicolumn{1}{c}{Acc} & \multicolumn{1}{c}{Psb} & \multicolumn{1}{c}{Sps} & \multicolumn{1}{c|}{Cost}        & \multicolumn{1}{c}{Acc} & \multicolumn{1}{c}{Psb} & \multicolumn{1}{c}{Sps} & \multicolumn{1}{c|}{Cost}        & \multicolumn{1}{c}{Acc} & \multicolumn{1}{c}{Psb} & \multicolumn{1}{c}{Sps} & \multicolumn{1}{c|}{Cost}        \\ \cline{2-13} 
\textbf{(RF) L-CR}  & \textcolor{RoyalBlue}{0.98} / \textcolor{OrangeRed}{0.93}             & \textcolor{RoyalBlue}{0.90} / \textcolor{OrangeRed}{0.92}             & \textcolor{RoyalBlue}{2.00} / \textcolor{OrangeRed}{3.00}             & \multicolumn{1}{l|}{\textcolor{RoyalBlue}{0.65} / \textcolor{OrangeRed}{0.99}} & \textcolor{RoyalBlue}{1.00} / \textcolor{OrangeRed}{1.00}             & \textcolor{RoyalBlue}{0.98} / \textcolor{OrangeRed}{0.89}             & \textcolor{RoyalBlue}{3.34} / \textcolor{OrangeRed}{3.74}             & \multicolumn{1}{l|}{\textcolor{RoyalBlue}{1.34} / \textcolor{OrangeRed}{1.63}} & \textcolor{RoyalBlue}{0.95} / \textcolor{OrangeRed}{0.97}             & \textcolor{RoyalBlue}{0.98} / \textcolor{OrangeRed}{0.98}             & \textcolor{RoyalBlue}{4.00} / \textcolor{OrangeRed}{5.00}             & \multicolumn{1}{l|}{\textcolor{RoyalBlue}{0.76} / \textcolor{OrangeRed}{1.16}} \\
\textbf{(RF) R-CR}  & \textcolor{RoyalBlue}{0.90} / \textcolor{OrangeRed}{0.90}             & \textcolor{RoyalBlue}{0.83} / \textcolor{OrangeRed}{0.88}             & \textcolor{RoyalBlue}{2.00} / \textcolor{OrangeRed}{3.00}             & \multicolumn{1}{l|}{\textcolor{RoyalBlue}{0.92} / \textcolor{OrangeRed}{0.99}} & \textcolor{RoyalBlue}{0.92} / \textcolor{OrangeRed}{0.80}             & \textcolor{RoyalBlue}{0.97} / \textcolor{OrangeRed}{0.89}             & \textcolor{RoyalBlue}{3.72} / \textcolor{OrangeRed}{3.98}             & \multicolumn{1}{l|}{\textcolor{RoyalBlue}{1.53} / \textcolor{OrangeRed}{1.67}} & \textcolor{RoyalBlue}{0.93} / \textcolor{OrangeRed}{0.97}             & \textcolor{RoyalBlue}{0.96} / \textcolor{OrangeRed}{0.97}             & \textcolor{RoyalBlue}{7.00} / \textcolor{OrangeRed}{7.00}             & \multicolumn{1}{l|}{\textcolor{RoyalBlue}{1.31} / \textcolor{OrangeRed}{1.33}} \\
\textbf{(RF) AReS}  & \textcolor{RoyalBlue}{0.72} / \textcolor{OrangeRed}{0.15}             & \textcolor{RoyalBlue}{0.39} / \textcolor{OrangeRed}{0.77}             & \textcolor{RoyalBlue}{1.84} / \textcolor{OrangeRed}{1.30}             & \multicolumn{1}{l|}{\textcolor{RoyalBlue}{0.34} / \textcolor{OrangeRed}{0.44}} & \textcolor{RoyalBlue}{0.72} / \textcolor{OrangeRed}{1.00}             & \textcolor{RoyalBlue}{0.85} / \textcolor{OrangeRed}{0.83}             & \textcolor{RoyalBlue}{1.09} / \textcolor{OrangeRed}{1.00}             & \multicolumn{1}{l|}{\textcolor{RoyalBlue}{0.28} / \textcolor{OrangeRed}{0.39}} & \textcolor{RoyalBlue}{0.92} / \textcolor{OrangeRed}{1.00}             & \textcolor{RoyalBlue}{0.84} / \textcolor{OrangeRed}{0.89}             & \textcolor{RoyalBlue}{1.00} / \textcolor{OrangeRed}{1.00}             & \multicolumn{1}{l|}{\textcolor{RoyalBlue}{0.25} / \textcolor{OrangeRed}{0.30}} \\
\textbf{(RF) Focus} & \textcolor{RoyalBlue}{0.00} / \textcolor{OrangeRed}{1.00}             & \textcolor{RoyalBlue}{0.00} / \textcolor{OrangeRed}{0.00}             & \textcolor{RoyalBlue}{8.61} / \textcolor{OrangeRed}{7.47}             & \multicolumn{1}{l|}{\textcolor{RoyalBlue}{2.50} / \textcolor{OrangeRed}{1.85}} & \textcolor{RoyalBlue}{0.00} / \textcolor{OrangeRed}{1.00}             & \textcolor{RoyalBlue}{0.00} / \textcolor{OrangeRed}{0.00}             & \textcolor{RoyalBlue}{8.00} / \textcolor{OrangeRed}{8.00}             & \multicolumn{1}{l|}{\textcolor{RoyalBlue}{2.21} / \textcolor{OrangeRed}{3.10}} & \textcolor{RoyalBlue}{0.00} / \textcolor{OrangeRed}{1.00}             & \textcolor{RoyalBlue}{0.00} / \textcolor{OrangeRed}{0.00}             & \textcolor{RoyalBlue}{17.0} / \textcolor{OrangeRed}{17.0}             & \multicolumn{1}{l|}{\textcolor{RoyalBlue}{3.28} / \textcolor{OrangeRed}{3.19}} \\
\textbf{(RF) FTW}   & \textcolor{RoyalBlue}{N/A} / \textcolor{OrangeRed}{1.00}              & \textcolor{RoyalBlue}{N/A} / \textcolor{OrangeRed}{0.22}              & \textcolor{RoyalBlue}{N/A} / \textcolor{OrangeRed}{3.00}              & \multicolumn{1}{l|}{\textcolor{RoyalBlue}{N/A} / \textcolor{OrangeRed}{0.97}}  & \textcolor{RoyalBlue}{N/A} / \textcolor{OrangeRed}{1.00}              & \textcolor{RoyalBlue}{N/A} / \textcolor{OrangeRed}{0.82}              & \textcolor{RoyalBlue}{N/A} / \textcolor{OrangeRed}{3.76}              & \multicolumn{1}{l|}{\textcolor{RoyalBlue}{N/A} / \textcolor{OrangeRed}{0.94}}  & \textcolor{RoyalBlue}{N/A} / \textcolor{OrangeRed}{1.00}              & \textcolor{RoyalBlue}{N/A} / \textcolor{OrangeRed}{0.76}              & \textcolor{RoyalBlue}{N/A} / \textcolor{OrangeRed}{3.92}              & \multicolumn{1}{l|}{\textcolor{RoyalBlue}{N/A} / \textcolor{OrangeRed}{0.41}}  \\
\multicolumn{13}{l}{}                                                                                                                                                                                                                                                                                                                                                  \\
\textbf{(NN) L-CR}  & \textcolor{RoyalBlue}{0.95} / \textcolor{OrangeRed}{0.88}             & \textcolor{RoyalBlue}{0.87} / \textcolor{OrangeRed}{0.79}             & \textcolor{RoyalBlue}{2.64} / \textcolor{OrangeRed}{3.25}             & \multicolumn{1}{l|}{\textcolor{RoyalBlue}{0.83} / \textcolor{OrangeRed}{0.86}} & \textcolor{RoyalBlue}{1.00} / \textcolor{OrangeRed}{1.00}             & \textcolor{RoyalBlue}{0.98} / \textcolor{OrangeRed}{0.89}             & \textcolor{RoyalBlue}{3.34} / \textcolor{OrangeRed}{3.71}             & \multicolumn{1}{l|}{\textcolor{RoyalBlue}{1.32} / \textcolor{OrangeRed}{1.63}} & \textcolor{RoyalBlue}{0.97} / \textcolor{OrangeRed}{0.96}             & \textcolor{RoyalBlue}{0.98} / \textcolor{OrangeRed}{0.97}             & \textcolor{RoyalBlue}{4.53} / \textcolor{OrangeRed}{5.59}             & \multicolumn{1}{l|}{\textcolor{RoyalBlue}{0.85} / \textcolor{OrangeRed}{1.08}} \\
\textbf{(NN) R-CR}  & \textcolor{RoyalBlue}{0.98} / \textcolor{OrangeRed}{0.88}             & \textcolor{RoyalBlue}{0.82} / \textcolor{OrangeRed}{0.91}             & \textcolor{RoyalBlue}{3.36} / \textcolor{OrangeRed}{3.29}             & \multicolumn{1}{l|}{\textcolor{RoyalBlue}{1.21} / \textcolor{OrangeRed}{1.00}} & \textcolor{RoyalBlue}{0.92} / \textcolor{OrangeRed}{0.78}             & \textcolor{RoyalBlue}{0.97} / \textcolor{OrangeRed}{0.89}             & \textcolor{RoyalBlue}{3.72} / \textcolor{OrangeRed}{3.93}             & \multicolumn{1}{l|}{\textcolor{RoyalBlue}{1.53} / \textcolor{OrangeRed}{1.67}} & \textcolor{RoyalBlue}{0.90} / \textcolor{OrangeRed}{0.93}             & \textcolor{RoyalBlue}{0.99} / \textcolor{OrangeRed}{0.98}             & \textcolor{RoyalBlue}{7.68} / \textcolor{OrangeRed}{7.76}             & \multicolumn{1}{l|}{\textcolor{RoyalBlue}{1.26} / \textcolor{OrangeRed}{1.39}} \\
\textbf{(NN) AReS}  & \textcolor{RoyalBlue}{0.92} / \textcolor{OrangeRed}{0.33}             & \textcolor{RoyalBlue}{0.83} / \textcolor{OrangeRed}{0.62}             & \textcolor{RoyalBlue}{1.18} / \textcolor{OrangeRed}{1.02}             & \multicolumn{1}{l|}{\textcolor{RoyalBlue}{0.52} / \textcolor{OrangeRed}{0.47}} & \textcolor{RoyalBlue}{1.00} / \textcolor{OrangeRed}{1.00}             & \textcolor{RoyalBlue}{0.87} / \textcolor{OrangeRed}{0.60}             & \textcolor{RoyalBlue}{1.00} / \textcolor{OrangeRed}{1.00}             & \multicolumn{1}{l|}{\textcolor{RoyalBlue}{0.29} / \textcolor{OrangeRed}{0.48}} & \textcolor{RoyalBlue}{0.99} / \textcolor{OrangeRed}{0.00}             & \textcolor{RoyalBlue}{0.81} / \textcolor{OrangeRed}{0.86}             & \textcolor{RoyalBlue}{1.00} / \textcolor{OrangeRed}{1.00}             & \multicolumn{1}{l|}{\textcolor{RoyalBlue}{0.26} / \textcolor{OrangeRed}{0.18}} \\
\textbf{(NN) CHVAE} & \textcolor{RoyalBlue}{N/A} / \textcolor{OrangeRed}{1.00}              & \textcolor{RoyalBlue}{N/A} / \textcolor{OrangeRed}{0.00}              & \textcolor{RoyalBlue}{N/A} / \textcolor{OrangeRed}{8.00}              & \multicolumn{1}{l|}{\textcolor{RoyalBlue}{N/A} / \textcolor{OrangeRed}{1.91}}  & \textcolor{RoyalBlue}{0.00} / \textcolor{OrangeRed}{1.00}             & \textcolor{RoyalBlue}{0.00} / \textcolor{OrangeRed}{0.00}             & \textcolor{RoyalBlue}{8.00} / \textcolor{OrangeRed}{8.00}             & \multicolumn{1}{l|}{\textcolor{RoyalBlue}{2.21} / \textcolor{OrangeRed}{3.10}} & \textcolor{RoyalBlue}{N/A} / \textcolor{OrangeRed}{1.00}              & \textcolor{RoyalBlue}{N/A} / \textcolor{OrangeRed}{0.00}              & \textcolor{RoyalBlue}{N/A} / \textcolor{OrangeRed}{17.0}              & \multicolumn{1}{l|}{\textcolor{RoyalBlue}{N/A} / \textcolor{OrangeRed}{3.68}}  \\
\end{tabular}
\end{table*}

To demonstrate the performance of our framework, we conduct two experiments on real-world datasets. In the first experiment, we showcase the utility of the Local Counterfactual Rules for explaining a regression model. In the second experiment, we compare our approaches with baseline methods in the context of classification problems. We compare the methods only in classification problems, as all prior works do not deal with regression problems. We use the library \emph{CARLA} \cite{pawelczyk2021carla} designed to benchmark counterfactual explanation methods across different data sets and machine learning models. 

We use the default settings of Random Forest and Neural Networks of \emph{CARLA} as query models. We compute the recourses given by our main competitor, AReS \citep{rawal2020beyond}, which performs an exhaustive search for finding global counterfactual rules. We use the implementation of \cite{cet4} that adapts AReS for returning counterfactual samples instead of rules. In cases where the Random Forest is the query model, we add two popular CF methods for tree-based models, FOCUS \cite{lucic2022focus} and FeatureTweaking \cite{tolomei2017interpretable} computed by \emph{CARLA}. We add CHVAE \cite{pawelczyk2020learning}, a CF method based on autoencoder for Neural Network experiments to enhance plausibility. In all experiments, we split our dataset into train ($75\%$) - test ($25\%$). We learn a model $f$, which could be either a Random Forest (RF) or Neural Network (NN), on the train set, which is the model we want to explain. We learn $f$'s predictions on the train set with an RF \textit{(estimators=20, max depth=10)} that will be used to generate the Counterfactual Rules (CR) with $\pi=0.9$. The parameters used for AReS are \textit{max rules=8, bins=10}, and the default parameters set by \emph{CARLA} for the other. For detailed parameter descriptions, see Appendix  (\ref{sup:additional_exp_paramaters}).

We evaluate the methods on the test set using four metrics. The first, \textit{Accuracy},  quantifies the average rate at which the recommended actions from each method successfully change the prediction to the desired outcome. The second, \textit{Plausibility}, measures the average number of counterfactual samples that are classified as inliers by an Isolation Forest trained on the training set. The third, \textit{Sparsity}, measures the average number of features that have been changed. The fourth metric is the \emph{Cost}, as defined in \cite{rawal2020beyond}. For categorical variables, any change is counted as a change of magnitude 1. The continuous features are converted into ordinal features by binning the feature values. In this case, going from one bin to the next immediate bin corresponds to a change of magnitude 1.

\textbf{Local counterfactual rules for regression.} We apply our approach to the California House Price dataset (n=20640, p=8) \citep{california_data}, which contains information about each district such as income, population, and location, and the goal is to predict the median house value of each district. To demonstrate the effectiveness of our Local CR method, we focus on a subset of the test set consisting of $1566$ houses with prices lower than $100k$. We aim to find recourse that would increase house prices, bringing them within the target range $\YSt=[200k, 250k]$. For each instance $\x$, we compute the Minimal Divergent Explanation $S$, the Local CR $C_S(\x, [200k, 250k])$, and generate a counterfactual sample using the Simulated Annealing technique described earlier. We succeed in changing the decision for all observations, achieving $\textit{Accuracy} = 100\%$. Furthermore, the majority of counterfactual samples passed the outlier test, with a $\textit{Plausibility}$ score of 0.92. Additionally, our Local CR method achieves high sparsity, with $\textit{Sparsity} = 4.45.$

For instance, the Local CR for the observation $\x =$ [Longitude=-118.2, latitude=33.8, housing median age=26, total rooms=703,
total bedrooms=202, population=757, households=212, median income=2.52] is $C_S(\x, [200k, 250k]) =$ [total room $\in [2132, 3546],$ total bedrooms $\in [214, 491]$] with probability 0.97. This means that if the total number of rooms and total bedrooms satisfy the conditions in $C_S(\x, [200k, 250k])$, and the remaining features of $\x$ are fixed, then the probability that the price falls within the target set $\YSt=[200k, 250k]$ is 0.97. This makes sense as increasing the number of rooms and bedrooms in the district will certainly increase the price. 

\textbf{Comparisons of Local and Regional CR with baseline methods.} We evaluate our framework on three real-world datasets: Compas (n=6172, p=12) \citep{washington2018argue} is used to predict criminal recidivism, Diabetes (n=768, p=8) \citep{diabetes} aims to predict whether a patient has diabetes or not, and Nhanesi \citep{nhanes} which also aims to predict a disease. Our evaluation reveals that AReS is highly sensitive to the number of bins and the maximal number of rules or actions, as previously noted by \citep{globalce}. Poor parameterization can result in completely ineffective recourses. Furthermore, these methods require separate models for each target class, while our framework only requires a single RF with good precision.

Table \ref{tab:results} shows the results of each method across all datasets and models (RF, NN). The results for the RF are in the first five rows, and those below are for the NN. In each cell, we have two values, the blue corresponding to the metric value for the positive class (label=1) and the red for the negative class (label=0). Table \ref{tab:results} demonstrates that the Local and Regional CR methods achieve high accuracy in changing decisions on all datasets, surpassing baseline methods in almost all experiments. Furthermore, the baselines struggle to change both the positive and negative classes simultaneously, e.g., ARes has \textit{Acc}=0.72 in the positive class, and 0.15 for the negative class on (RF) - Compas or when they have a good \textit{Acc}, the CE are not plausible. For instance, FOCUS has \textit{Acc}=1 for the negative class and \textit{Psb}=0 in this experiment, meaning that all the counterfactual samples are outliers. These trends are consistently observed across most experiments. We also noted that Focus and FTW fail completely to change the positive class, and CHVAE often does not even propose a CF for the positive class. This could be attributed to these methods being more sensitive to imbalances in the data. Our method is the only one that works consistently with the positive and negative classes. Interestingly, baseline methods exhibit improved performance on the NN model compared to the RF model.

Regarding plausibility, our method generally outperforms others, albeit at a higher cost. This is not surprising, as all baseline methods optimise this criterion, unlike ours. In the future, we intend to add this criterion to the CF generation process (Section \ref{sec:sampling_ce}).


\textbf{Noisy responses robustness of Local CR:} To assess the robustness of our approach against noisy responses, we conduct an experiment inspired by \cite{himanoisycounterfactuals}. We normalized the datasets so that $\X \in [0, 1]^p$ and added small Gaussian noises $\epsilon$ to the prescribed recourses, with $\epsilon \sim \mathcal{N}(0, \sigma^2)$, where $\sigma^2$ took values of $0.01, 0.025, 0.05$. We compute the \textit{Stability}, which is the fraction of unseen instances where the action and perturbed action lead to the same output, for the Compas and Diabetes datasets. We used the simulated annealing approach of Section \ref{sec:sampling_ce}  with the Local CR to generate the actions. The Stability metrics for the different noise levels were $0.98, 0.98, 0.98$ for Compas and $0.96, 0.97, 0.96$ for Diabetes.

In summary, our CR approach is easier to train, and provides more accurate and plausible rules than the baseline methods. Furthermore, our resulting CE is robust against noisy responses.

\section{Conclusion}
We propose a novel approach that formulates CE as \textit{Counterfactual Rules}. These rules are simple policies that can change the decision of an individual or sub-population with a high probability. Our method is designed to learn robust, plausible, and sparse adversarial regions that indicate where observations should be moved to satisfy a desired outcome. Random Forests are central to our approach, as they provide consistent estimates of the probabilities of interest and naturally give rise to the counterfactual rules we seek. This also allows us to handle regression problems and continuous features, making our method applicable to a wide range of data sets where tree-based models perform well, such as tabular data \citep{grinsztajn2022tree}. An interesting avenue to explore would be to incorporate the $l_1$ cost into our approach. Currently, our method aims to minimize the $l_0$ distance between the query $\x^{obs}$ and the counterfactual $\x^{CF}$ by altering as few features as possible. However, deriving a counterfactual observation within a counterfactual rule that minimizes the $l_1$ cost is straightforward with an explicit solution. Given the counterfactual rules (hyperrectangles), represented as a box $(\textbf{l}, \textbf{r})$, with $\textbf{l}, \textbf{r} \in \mathbb{R}^p$, the following optimization problem $\x^{CF} = argmin_{\x} d(\x, \x^{obs})$ such that $\textbf{l} \leq \x \leq \textbf{r}$ has a closed form solution when the distance is the $l_1$ or $l_2$ norm. The solution is $\x^{CF} = \max(\textbf{l}, \min(\textbf{r}, \x^{obs}))$ elementwise \citep{carreira2021counterfactual}. In future work, we will incorporate the $l_1$ constraint and assess the effectiveness of our approach in terms of cost relative to other methods.

\bibliography{example_paper}

\begin{thebibliography}{55}
\providecommand{\natexlab}[1]{#1}
\providecommand{\url}[1]{\texttt{#1}}
\expandafter\ifx\csname urlstyle\endcsname\relax
  \providecommand{\doi}[1]{doi: #1}\else
  \providecommand{\doi}{doi: \begingroup \urlstyle{rm}\Url}\fi

\bibitem[Amoukou \& Brunel(2021)Amoukou and Brunel]{amoukou2021consistent}
Amoukou, S.~I. and Brunel, N.~J.
\newblock Consistent sufficient explanations and minimal local rules for explaining regression and classification models.
\newblock \emph{arXiv preprint arXiv:2111.04658}, 2021.

\bibitem[Basu et~al.(2018)Basu, Kumbier, Brown, and Yu]{basu2018iterative}
Basu, S., Kumbier, K., Brown, J.~B., and Yu, B.
\newblock Iterative random forests to discover predictive and stable high-order interactions.
\newblock \emph{Proceedings of the National Academy of Sciences}, 115\penalty0 (8):\penalty0 1943--1948, 2018.

\bibitem[B{\'e}nard et~al.(2021{\natexlab{a}})B{\'e}nard, Biau, Da~Veiga, and Scornet]{benard2021shaff}
B{\'e}nard, C., Biau, G., Da~Veiga, S., and Scornet, E.
\newblock Shaff: Fast and consistent shapley effect estimates via random forests.
\newblock \emph{arXiv preprint arXiv:2105.11724}, 2021{\natexlab{a}}.

\bibitem[B{\'e}nard et~al.(2021{\natexlab{b}})B{\'e}nard, Biau, Veiga, and Scornet]{benard2021interpretable}
B{\'e}nard, C., Biau, G., Veiga, S., and Scornet, E.
\newblock Interpretable random forests via rule extraction.
\newblock In \emph{International Conference on Artificial Intelligence and Statistics}, pp.\  937--945. PMLR, 2021{\natexlab{b}}.

\bibitem[B{\'e}nard et~al.(2021{\natexlab{c}})B{\'e}nard, Da~Veiga, and Scornet]{benard2021mda}
B{\'e}nard, C., Da~Veiga, S., and Scornet, E.
\newblock Mda for random forests: inconsistency, and a practical solution via the sobol-mda.
\newblock \emph{arXiv preprint arXiv:2102.13347}, 2021{\natexlab{c}}.

\bibitem[Biau \& Devroye(2010)Biau and Devroye]{biau_layered_2010}
Biau, G. and Devroye, L.
\newblock On the layered nearest neighbour estimate, the bagged nearest neighbour estimate and the random forest method in regression and classification.
\newblock \emph{Journal of Multivariate Analysis}, 101\penalty0 (10):\penalty0 2499--2518, 2010.
\newblock ISSN 0047-259X.
\newblock \doi{https://doi.org/10.1016/j.jmva.2010.06.019}.
\newblock URL \url{https://www.sciencedirect.com/science/article/pii/S0047259X10001387}.

\bibitem[Black et~al.(2020)Black, Yeom, and Fredrikson]{black2020fliptest}
Black, E., Yeom, S., and Fredrikson, M.
\newblock Fliptest: fairness testing via optimal transport.
\newblock In \emph{Proceedings of the 2020 Conference on Fairness, Accountability, and Transparency}, pp.\  111--121, 2020.

\bibitem[Breiman(2000)]{breiman2000some}
Breiman, L.
\newblock Some infinity theory for predictor ensembles.
\newblock Technical report, Citeseer, 2000.

\bibitem[Breiman et~al.(1984)Breiman, Friedman, Olshen, and Stone]{breiman1984classification}
Breiman, L., Friedman, J., Olshen, R., and Stone, C.
\newblock Classification and regression trees. wadsworth int.
\newblock \emph{Group}, 37\penalty0 (15):\penalty0 237--251, 1984.

\bibitem[Carreira-Perpi{\~n}{\'a}n \& Hada(2021)Carreira-Perpi{\~n}{\'a}n and Hada]{carreira2021counterfactual}
Carreira-Perpi{\~n}{\'a}n, M.~{\'A}. and Hada, S.~S.
\newblock Counterfactual explanations for oblique decision trees: Exact, efficient algorithms.
\newblock In \emph{Proceedings of the AAAI conference on artificial intelligence}, volume~35, pp.\  6903--6911, 2021.

\bibitem[CDC(1999-2022)]{nhanes}
CDC.
\newblock National health and nutrition examination survey, 1999-2022.
\newblock URL \url{https://wwwn.cdc.gov/Nchs/Nhanes/Default.aspx.}

\bibitem[Chou et~al.(2022)Chou, Moreira, Bruza, Ouyang, and Jorge]{CHOU202259}
Chou, Y.-L., Moreira, C., Bruza, P., Ouyang, C., and Jorge, J.
\newblock Counterfactuals and causability in explainable artificial intelligence: Theory, algorithms, and applications.
\newblock \emph{Information Fusion}, 81:\penalty0 59--83, 2022.
\newblock ISSN 1566-2535.
\newblock \doi{https://doi.org/10.1016/j.inffus.2021.11.003}.
\newblock URL \url{https://www.sciencedirect.com/science/article/pii/S1566253521002281}.

\bibitem[De~Lara et~al.(2021)De~Lara, Gonz{\'a}lez-Sanz, Asher, and Loubes]{de2021transport}
De~Lara, L., Gonz{\'a}lez-Sanz, A., Asher, N., and Loubes, J.-M.
\newblock Transport-based counterfactual models.
\newblock \emph{arXiv preprint arXiv:2108.13025}, 2021.

\bibitem[Du et~al.(2021)Du, Biau, Petit, and Porcher]{du2021wasserstein}
Du, Q., Biau, G., Petit, F., and Porcher, R.
\newblock Wasserstein random forests and applications in heterogeneous treatment effects.
\newblock In \emph{International Conference on Artificial Intelligence and Statistics}, pp.\  1729--1737. PMLR, 2021.

\bibitem[FICO(2018)]{helocdata}
FICO.
\newblock Fico. explainable machine learning challenge, 2018.
\newblock URL \url{https://community.fico.com/ s/explainable-machine-learning-challenge.}

\bibitem[Geurts et~al.(2006)Geurts, Ernst, and Wehenkel]{geurts2006extremely}
Geurts, P., Ernst, D., and Wehenkel, L.
\newblock Extremely randomized trees.
\newblock \emph{Machine learning}, 63:\penalty0 3--42, 2006.

\bibitem[Grathwohl et~al.(2020)Grathwohl, Wang, Jacobsen, Duvenaud, Norouzi, and Swersky]{ebmduvenaud}
Grathwohl, W., Wang, K.-C., Jacobsen, J.-H., Duvenaud, D., Norouzi, M., and Swersky, K.
\newblock Your classifier is secretly an energy based model and you should treat it like one.
\newblock In \emph{International Conference on Learning Representations}, 2020.

\bibitem[Grinsztajn et~al.(2022)Grinsztajn, Oyallon, and Varoquaux]{grinsztajn2022tree}
Grinsztajn, L., Oyallon, E., and Varoquaux, G.
\newblock Why do tree-based models still outperform deep learning on typical tabular data?
\newblock In \emph{Thirty-sixth Conference on Neural Information Processing Systems Datasets and Benchmarks Track}, 2022.

\bibitem[Guilmeau et~al.(2021)Guilmeau, Chouzenoux, and Elvira]{review_simulated_annealing}
Guilmeau, T., Chouzenoux, E., and Elvira, V.
\newblock Simulated annealing: a review and a new scheme.
\newblock pp.\  101--105, 07 2021.
\newblock \doi{10.1109/SSP49050.2021.9513782}.

\bibitem[Ishwaran et~al.(2008)Ishwaran, Kogalur, Blackstone, and Lauer]{ishwaran2008random}
Ishwaran, H., Kogalur, U.~B., Blackstone, E.~H., and Lauer, M.~S.
\newblock Random survival forests.
\newblock \emph{The annals of applied statistics}, 2\penalty0 (3):\penalty0 841--860, 2008.

\bibitem[Jocteur et~al.(2023)Jocteur, Maume-Deschamps, and Ribereau]{jocteur2023heterogeneous}
Jocteur, B.-A., Maume-Deschamps, V., and Ribereau, P.
\newblock Heterogeneous treatment effect based random forest: Hterf.
\newblock 2023.

\bibitem[Kaggle(2016)]{diabetes}
Kaggle.
\newblock Pima indians diabetes database, 2016.
\newblock URL \url{https://www.kaggle.com/datasets/uciml/pima-indians-diabetes-database}.

\bibitem[Kanamori et~al.(2022)Kanamori, Takagi, Kobayashi, and Ike]{cet4}
Kanamori, K., Takagi, T., Kobayashi, K., and Ike, Y.
\newblock Counterfactual explanation trees: Transparent and consistent actionable recourse with decision trees.
\newblock In \emph{Proceedings of The 25th International Conference on Artificial Intelligence and Statistics, PMLR 151:1846-1870}, 2022.

\bibitem[Karimi et~al.(2020{\natexlab{a}})Karimi, Barthe, Sch{\"{o}}lkopf, and Valera]{survey_counterfactual}
Karimi, A., Barthe, G., Sch{\"{o}}lkopf, B., and Valera, I.
\newblock A survey of algorithmic recourse: definitions, formulations, solutions, and prospects.
\newblock \emph{CoRR}, abs/2010.04050, 2020{\natexlab{a}}.
\newblock URL \url{https://arxiv.org/abs/2010.04050}.

\bibitem[Karimi et~al.(2020{\natexlab{b}})Karimi, Barthe, Balle, and Valera]{Karimi2020ModelAgnosticCE}
Karimi, A.-H., Barthe, G., Balle, B., and Valera, I.
\newblock Model-agnostic counterfactual explanations for consequential decisions.
\newblock \emph{ArXiv}, abs/1905.11190, 2020{\natexlab{b}}.

\bibitem[{Kelley Pace} \& Barry(1997){Kelley Pace} and Barry]{california_data}
{Kelley Pace}, R. and Barry, R.
\newblock Sparse spatial autoregressions.
\newblock \emph{Statistics, Probability Letters}, 33\penalty0 (3):\penalty0 291--297, 1997.
\newblock ISSN 0167-7152.
\newblock \doi{https://doi.org/10.1016/S0167-7152(96)00140-X}.
\newblock URL \url{https://www.sciencedirect.com/science/article/pii/S016771529600140X}.

\bibitem[Lakkaraju et~al.(2022)Lakkaraju, Slack, Chen, Tan, and Singh]{rethinkinxai}
Lakkaraju, H., Slack, D., Chen, Y., Tan, C., and Singh, S.
\newblock Rethinking explainability as a dialogue: {A} practitioner's perspective.
\newblock \emph{CoRR}, abs/2202.01875, 2022.
\newblock URL \url{https://arxiv.org/abs/2202.01875}.

\bibitem[Lecun et~al.(2006)Lecun, Chopra, and Hadsell]{yanebm}
Lecun, Y., Chopra, S., and Hadsell, R.
\newblock \emph{A tutorial on energy-based learning}.
\newblock 01 2006.

\bibitem[Ley et~al.(2022)Ley, Mishra, and Magazzeni]{globalce}
Ley, D., Mishra, S., and Magazzeni, D.
\newblock Global counterfactual explanations: Investigations, implementations and improvements, 2022.
\newblock URL \url{https://arxiv.org/abs/2204.06917}.

\bibitem[Lin \& Jeon(2006)Lin and Jeon]{lin2006random}
Lin, Y. and Jeon, Y.
\newblock Random forests and adaptive nearest neighbors.
\newblock \emph{Journal of the American Statistical Association}, 101\penalty0 (474):\penalty0 578--590, 2006.

\bibitem[Liu et~al.(2008)Liu, Ting, and Zhou]{liu2008isolation}
Liu, F.~T., Ting, K.~M., and Zhou, Z.-H.
\newblock Isolation forest.
\newblock In \emph{2008 eighth ieee international conference on data mining}, pp.\  413--422. IEEE, 2008.

\bibitem[Loh(2011)]{Loh2011ClassificationAR}
Loh, W.-Y.
\newblock Classification and regression trees.
\newblock \emph{Wiley Interdisciplinary Reviews: Data Mining and Knowledge Discovery}, 1, 2011.

\bibitem[Looveren \& Klaise(2019)Looveren and Klaise]{prototype_basedce}
Looveren, A.~V. and Klaise, J.
\newblock Interpretable counterfactual explanations guided by prototypes.
\newblock \emph{CoRR}, abs/1907.02584, 2019.
\newblock URL \url{http://arxiv.org/abs/1907.02584}.

\bibitem[Lucic et~al.(2022)Lucic, Oosterhuis, Haned, and de~Rijke]{lucic2022focus}
Lucic, A., Oosterhuis, H., Haned, H., and de~Rijke, M.
\newblock Focus: Flexible optimizable counterfactual explanations for tree ensembles.
\newblock In \emph{Proceedings of the AAAI Conference on Artificial Intelligence}, volume~36, pp.\  5313--5322, 2022.

\bibitem[Lundberg et~al.(2020)Lundberg, Erion, Chen, DeGrave, Prutkin, Nair, Katz, Himmelfarb, Bansal, and Lee]{lundberg2020local2global}
Lundberg, S.~M., Erion, G., Chen, H., DeGrave, A., Prutkin, J.~M., Nair, B., Katz, R., Himmelfarb, J., Bansal, N., and Lee, S.-I.
\newblock From local explanations to global understanding with explainable ai for trees.
\newblock \emph{Nature Machine Intelligence}, 2\penalty0 (1):\penalty0 2522--5839, 2020.

\bibitem[Meinshausen \& Ridgeway(2006)Meinshausen and Ridgeway]{meinshausen2006quantile}
Meinshausen, N. and Ridgeway, G.
\newblock Quantile regression forests.
\newblock \emph{Journal of Machine Learning Research}, 7\penalty0 (6), 2006.

\bibitem[Molnar(2022)]{molnar2022}
Molnar, C.
\newblock \emph{Interpretable Machine Learning}.
\newblock 2 edition, 2022.
\newblock URL \url{https://christophm.github.io/interpretable-ml-book}.

\bibitem[Mothilal et~al.(2020)Mothilal, Sharma, and Tan]{dice}
Mothilal, R.~K., Sharma, A., and Tan, C.
\newblock Explaining machine learning classifiers through diverse counterfactual explanations.
\newblock In \emph{Proceedings of the 2020 Conference on Fairness, Accountability, and Transparency}, FAT* '20, pp.\  607–617, New York, NY, USA, 2020. Association for Computing Machinery.
\newblock ISBN 9781450369367.
\newblock \doi{10.1145/3351095.3372850}.
\newblock URL \url{https://doi.org/10.1145/3351095.3372850}.

\bibitem[Parmentier \& Vidal(2021)Parmentier and Vidal]{optimalce_vidal}
Parmentier, A. and Vidal, T.
\newblock Optimal counterfactual explanations in tree ensembles.
\newblock \emph{CoRR}, abs/2106.06631, 2021.
\newblock URL \url{https://arxiv.org/abs/2106.06631}.

\bibitem[{Patki} et~al.(2016){Patki}, {Wedge}, and {Veeramachaneni}]{sdv}
{Patki}, N., {Wedge}, R., and {Veeramachaneni}, K.
\newblock The synthetic data vault.
\newblock In \emph{2016 IEEE International Conference on Data Science and Advanced Analytics (DSAA)}, pp.\  399--410, Oct 2016.
\newblock \doi{10.1109/DSAA.2016.49}.

\bibitem[Pawelczyk et~al.(2020)Pawelczyk, Broelemann, and Kasneci]{pawelczyk2020learning}
Pawelczyk, M., Broelemann, K., and Kasneci, G.
\newblock Learning model-agnostic counterfactual explanations for tabular data.
\newblock In \emph{Proceedings of the web conference 2020}, pp.\  3126--3132, 2020.

\bibitem[Pawelczyk et~al.(2021)Pawelczyk, Bielawski, van~den Heuvel, Richter, and Kasneci]{pawelczyk2021carla}
Pawelczyk, M., Bielawski, S., van~den Heuvel, J., Richter, T., and Kasneci, G.
\newblock Carla: A python library to benchmark algorithmic recourse and counterfactual explanation algorithms, 2021.

\bibitem[Pawelczyk et~al.(2022)Pawelczyk, Datta, van-den Heuvel, Kasneci, and Lakkaraju]{himanoisycounterfactuals}
Pawelczyk, M., Datta, T., van-den Heuvel, J., Kasneci, G., and Lakkaraju, H.
\newblock Algorithmic recourse in the face of noisy human responses, 2022.
\newblock URL \url{https://arxiv.org/abs/2203.06768}.

\bibitem[Poyiadzi et~al.(2019)Poyiadzi, Sokol, Santos{-}Rodriguez, Bie, and Flach]{face_counterfactual}
Poyiadzi, R., Sokol, K., Santos{-}Rodriguez, R., Bie, T.~D., and Flach, P.~A.
\newblock {FACE:} feasible and actionable counterfactual explanations.
\newblock \emph{CoRR}, abs/1909.09369, 2019.
\newblock URL \url{http://arxiv.org/abs/1909.09369}.

\bibitem[Rawal \& Lakkaraju(2020)Rawal and Lakkaraju]{rawal2020beyond}
Rawal, K. and Lakkaraju, H.
\newblock Beyond individualized recourse: Interpretable and interactive summaries of actionable recourses.
\newblock \emph{Advances in Neural Information Processing Systems}, 33:\penalty0 12187--12198, 2020.

\bibitem[Ribeiro et~al.(2016)Ribeiro, Singh, and Guestrin]{ribeiro2016why}
Ribeiro, M.~T., Singh, S., and Guestrin, C.
\newblock " why should i trust you?" explaining the predictions of any classifier.
\newblock In \emph{Proceedings of the 22nd ACM SIGKDD international conference on knowledge discovery and data mining}, pp.\  1135--1144, 2016.

\bibitem[Russell(2019)]{diverce_ce}
Russell, C.
\newblock Efficient search for diverse coherent explanations.
\newblock In \emph{Proceedings of the Conference on Fairness, Accountability, and Transparency}, FAT* '19, pp.\  20–28, New York, NY, USA, 2019. Association for Computing Machinery.
\newblock ISBN 9781450361255.
\newblock \doi{10.1145/3287560.3287569}.
\newblock URL \url{https://doi.org/10.1145/3287560.3287569}.

\bibitem[Scornet(2016)]{scornet2016random}
Scornet, E.
\newblock Random forests and kernel methods.
\newblock \emph{IEEE Transactions on Information Theory}, 62\penalty0 (3):\penalty0 1485--1500, 2016.

\bibitem[Tolomei et~al.(2017)Tolomei, Silvestri, Haines, and Lalmas]{tolomei2017interpretable}
Tolomei, G., Silvestri, F., Haines, A., and Lalmas, M.
\newblock Interpretable predictions of tree-based ensembles via actionable feature tweaking.
\newblock In \emph{Proceedings of the 23rd ACM SIGKDD international conference on knowledge discovery and data mining}, pp.\  465--474, 2017.

\bibitem[Ustun et~al.(2019)Ustun, Spangher, and Liu]{Ustun2019ActionableRI}
Ustun, B., Spangher, A., and Liu, Y.
\newblock Actionable recourse in linear classification.
\newblock \emph{Proceedings of the Conference on Fairness, Accountability, and Transparency}, 2019.

\bibitem[Verma et~al.(2020)Verma, Dickerson, and Hines]{counterfactual_r1}
Verma, S., Dickerson, J.~P., and Hines, K.
\newblock Counterfactual explanations for machine learning: {A} review.
\newblock \emph{CoRR}, abs/2010.10596, 2020.
\newblock URL \url{https://arxiv.org/abs/2010.10596}.

\bibitem[Wachter et~al.(2017)Wachter, Mittelstadt, and Russell]{Wachter2017CounterfactualEW}
Wachter, S., Mittelstadt, B.~D., and Russell, C.
\newblock Counterfactual explanations without opening the black box: Automated decisions and the gdpr.
\newblock \emph{Cybersecurity}, 2017.

\bibitem[Wager \& Athey(2017)Wager and Athey]{wager2017estimation}
Wager, S. and Athey, S.
\newblock Estimation and inference of heterogeneous treatment effects using random forests, 2017.

\bibitem[Washington(2018)]{washington2018argue}
Washington, A.~L.
\newblock How to argue with an algorithm: Lessons from the compas-propublica debate.
\newblock \emph{Colo. Tech. LJ}, 17:\penalty0 131, 2018.

\bibitem[Xu et~al.(2019)Xu, Skoularidou, Cuesta-Infante, and Veeramachaneni]{modeling_td}
Xu, L., Skoularidou, M., Cuesta-Infante, A., and Veeramachaneni, K.
\newblock Modeling tabular data using conditional gan.
\newblock In \emph{NeurIPS}, 2019.

\end{thebibliography}
\bibliographystyle{icml2024}

\newpage
\appendix
\onecolumn
\section{Regional RF detailed} \label{sup:regional_forest}
In this section, we give a simple application of the Regional RF algorithm to better understand how it works. Recall that the Regional RF is a generalization of the RF's algorithm to give prediction even when we condition given a region, e.g., to estimate $E(f(\X) \; | \boldsymbol{X}_S \in C_S(\boldsymbol{x}), \boldsymbol{X}_{\bar{S}} = \boldsymbol{x}_{\bar{S}})$ with $C_{S}(\boldsymbol{x}) = \prod_{i=1}^{|S|} [a_i, b_i], a_i, b_i \in \bar{\mathbb{R}}$ a hyperrectangle. The algorithm works as follows: we drop the observations in the initial trees, if a split used variable $i \in \bar{S}$, a fixed value-based condition, we used the classic rules, i.e.,  if $x_i \leq t$, the observations go to the left children, otherwise the right children. However, if a split used variable $i \in S$, regional-based condition, we used the hyperrectangle $C_S(\boldsymbol{x}) = \prod_{i=1}^{|S|} [a_i, b_i]$. The observations are sent to the left children if $b_i \leq t$, right children if $a_i > t$ and if $t \in [a_i, b_i]$ the observations are sent both to the left and right children. 

To illustrate how it works, we use a two dimensional variables $\X = (X_0, X_1)\in \mathbb{R}^2$, a simple decision tree $f$ represented in Figure \ref{fig:tree_example}, and want to compute for $\x = [1.5, 1.9],$ $\mathbb{E}(f(\X) \;|\; X_1 \in [2, \; 3.5], X_0 = 1.5)$. We assume that $\mathbb{P}(X_1 \in [2, \; 3.5] \; | \; X_0 = 1.5) >0$ and denoted $T_1$ as the set of the values of the splits based on variables $X_1$ of the decision tree. One way of estimating this conditional mean is by using Monte Carlo sampling. Therefore, there are two cases : 

\begin{figure}[ht!]
    \centering
    \includegraphics[scale=0.45]{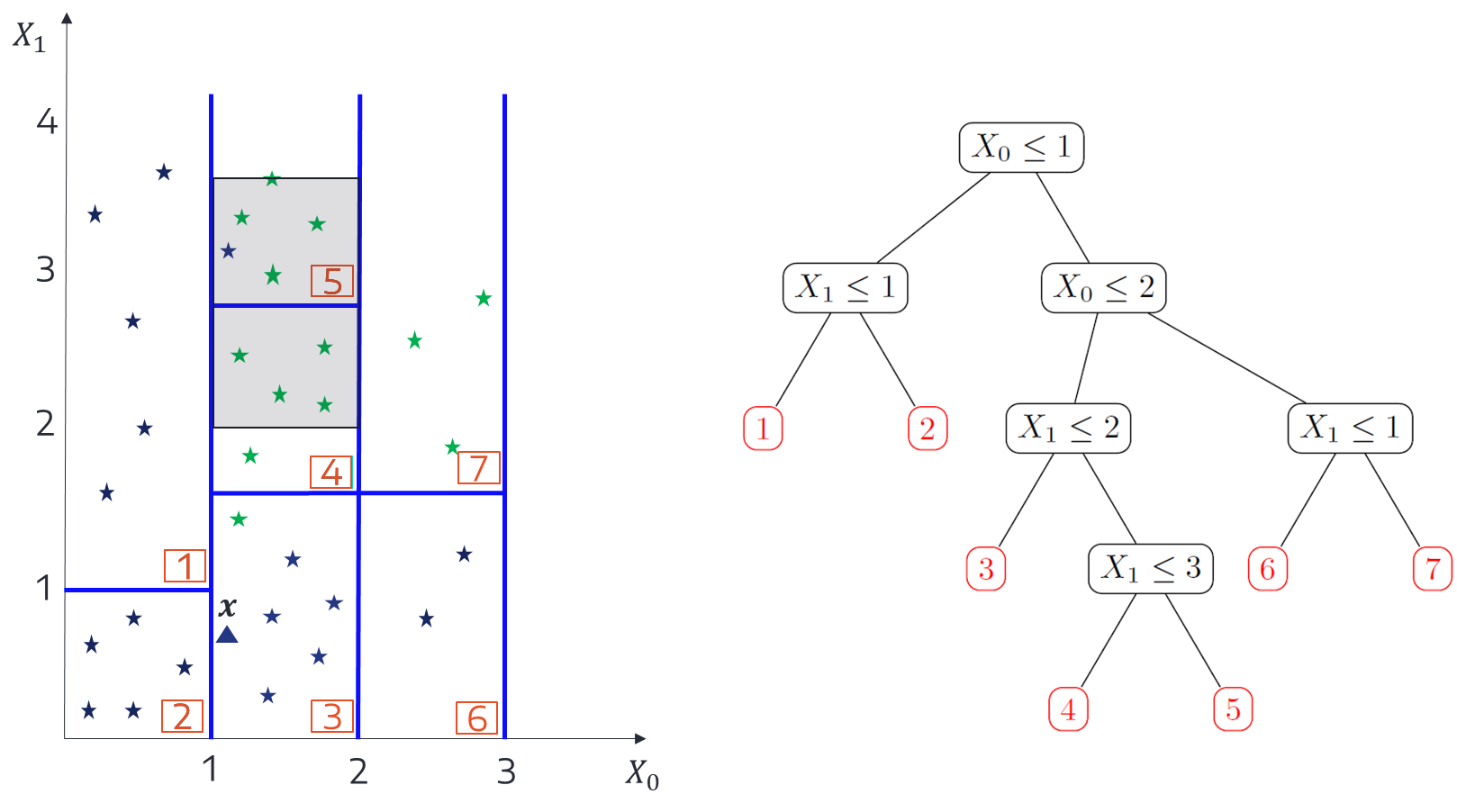}
    \caption{Representation of a simple decision tree (right Figure) and its associated partition (left Figure). The gray part in the partition corresponds to the region $[2, \; 3.5] \times [1, 2]$}
    \label{fig:tree_example}
\end{figure}
\begin{itemize}
    \item If $\forall t \in T_1,$ $t \leq 2$ or $t > 3.5$, then all the observations sampled s.t. $\Tilde{X}_i \sim P_{\X \; |\; X_1 \in [2, \; 3.5], X_0 = 1.5}$ follow the same path and fall in the same leaf. The Monte Carlo estimator of the decision tree $\mathbb{E}(f(\X) | X_1 \in [2, \; 3.5], X_0 = 1.5)$ is equal to the output of the Regional RF algorithm. 
    \begin{itemize}
        \item For instance, a special case of the case above is: if $\forall t \in T_1, t \leq 2$, and we sample using $P_{\X \; | \; X_1 \in [2, \; 3.5], X_0 = 1.5}$,  then all the observations go to the right children when they encounters a node using $X_1$ and fall in the same leaf. 
    \end{itemize}
    \item If $\exists t \in T_1$ and $t \in [2, \; 3.5]$, then the observations sampled s.t. $\Tilde{X}_i \sim P_{\X \; | \; X_1 \in [2, \; 3.5], X_0 = 1.5}$ can fall in multiple terminal leaf depending on if their coordinates $x_1$ is lower than $t$. Following our example,  if we generate samples using $P_{\X \; | \; X_1 \in [2, \; 3.5], X_0 = 1.5}$, the observations will fall in the gray region of Figure \ref{fig:tree_example}, and thus can fall in node 4 or 5. Therefore, the true estimate is: 
    \begin{multline} \label{fig:weighted_mean}
        \small  \mathbb{E}(f(\X) \; | \;X_1 \in [2, \; 3.5], X_0 = 1.5 ) = 
        \mathbb{P}(X_1 \leq 2.9\; | X_0=1.5)\times \mathbb{E}[f(\X)\;| \X \in L_4] \\ + \mathbb{P}(X_1 > 2.9\; | X_0=1.5)\times \mathbb{E}[f(\X)\; |\X \in L_5] 
    \end{multline}
\end{itemize}

Concerning the last case $(t \in [2, \; 3.5])$, we need to estimate the different probabilities $\mathbb{P}(X_1 \leq 2.9\; | \; X_0=1.5), \mathbb{P}(X_1 > 2.9\; | \;X_0=1.5)$ to compute $\mathbb{E}(f(\X) \;|\; X_1 \in [2, \; 3.5], X_0 = 1.5 )$, but these probabilities are difficult to estimate in practice. However, we argue that we can ignore these splits, and thus do no need to fragment the query region using the leaves of the tree. Indeed, as we are no longer interest in a point estimate but regional (population mean) we do not need to go to the level of the leaves. We propose to ignore the splits of the leaves that divide the query region.  For instance, the leaves 4 and 5 split the region $[2, \; 3.5]$ in two cells, by ignoring these splits we estimate the mean of the gray region by taking the average output of the leaves 4 and 5 instead of computing the mean weighted by the probabilities as in Equation (\ref{fig:weighted_mean}). Roughly, it consists to follow the classic rules of a decision tree (if the region is above or below a split) and ignore the splits that are in the query region, i.e., we average the output of all the leaves that are compatible with the condition $X_1 \in [2, \; 3.5], X_0 = 1.5$. 
We think it leads to a better estimation for two reasons. First, we observe that the case where $t$ is in the region and thus divides the query region does not occur often. Moreover, the leaves of the trees are very small in practice, and taking the mean of observations that fall into the union of leaves that belong to the query region is more reasonable than computing the weighted mean and thus trying to estimate the different probabilities $\mathbb{P}(X_1 \leq 2.9\; | X_0=1.5), \mathbb{P}(X_1 > 2.9\; | X_0=1.5)$.

\section{Sampling Recourses from counterfactual rules} \label{sec:algo}
\begin{algorithm}[ht]
\caption{Simulated Annealing to generate counterfactual samples using the Counterfactual Rules}
    \SetKwInOut{KwIn}{Input}
    \SetKwInOut{KwOut}{Output}

    \KwIn{Observation $\x$, Divergent Explanation $S$, counterfactual rule $C_S(\x, \YSt)$,  $\mathcal{D}_n$ training data set, number of iterations $maxIter$,  temperature $T$, cooling rate $r$}
    \KwOut{Inlier sample $\x^{best}$}
\begin{algorithmic}[1] 
\STATE Set  $\x^{current} \gets \x$, and $\x^{best} \gets \x$
\FOR{$j \in S$}
    \STATE $x^{current}_j \gets \text{sample uniformly from the set }\{X_{i, j}:  \X_i \in \mathcal{D}_n \text{  and  } \X_{i, S} \in C_S(\x, \YSt) \}$ \tcc*[r]{Generate $\x^{current} = (\boldsymbol{z}_S, \x_{\bar{S}})$ with $\boldsymbol{z}_S$ drawn using $\boldsymbol{z}_S \sim \prod_{i \in S} \hat{P}_{X_j \; | \X_S \in C_S(\x, \YSt)}$.} 
    \STATE  $x^{best}_j \gets x^{current}_j$\tcc*[r]{Initialize $x^{best}$}
\ENDFOR
\FOR{$it$ from $1$ to $maxIter$}
    \STATE $\x^{new} \gets \x^{current}$
    \STATE $S^\prime \gets \text{sample uniformly from the set } S$
    \FOR{$j$ in $S^\prime$}
        \STATE $x^{new}_j \gets \text{sample uniformly from the set }\{X_{i, j}:  \X_i \in \mathcal{D}_n \text{  and  } \X_{i, S} \in C_{S^\prime}(\x, \YSt) \}$  
    \ENDFOR
    \STATE Compute the Outlier score difference $\Delta O$ between $\x^{new}$ and $\x^{current}$
    \IF{$\Delta O < 0$ \textbf{or} $exp(-\Delta O / T) > random(0,1)$} 
        \STATE Set $\x^{current} \gets \x^{new}$
    \ENDIF
    \IF{Outlier score of $\x^{best} < $ Outlier score of $\x^{current}$}
        \STATE Set $\x^{best} \gets \x^{current}$
    \ENDIF
    \STATE Decrease $T$ by $T = T * r$
\ENDFOR
\STATE \textbf{return} $\x^{best}$
\end{algorithmic}
 \label{fig:simulated_ce_generation}
\end{algorithm}

\section{Additional experiments} \label{sup:additional_exp_paramaters}
In table \ref{tab:add_exp}, we compare the \textit{Accuracy} (Acc), \textit{Plausibility} (Psb), and \textit{Sparsity} (Sprs) of the different methods on additonal real-world datasets: FICO \citep{helocdata}, NHANESI \citep{nhanes}. 

We observe that the L-CR, and R-CR outperform the baseline methods by a large margin on \textit{Accuracy} and \textit{Plausibility}. The baseline methods still struggle to change at the same time the positive and negative class. AReS and CET give better sparsity, but their counterfactual samples are less plausible than the ones generated by the CR.

\begin{table}[ht!]
\small 
\caption{Results of the \textit{Accuracy} (Acc), \textit{Plausibility}, and \textit{Sparsity} (Sprs) of the different methods. We compute each metric according to the positive (Pos) and negative (Neg) class.}
\label{tab:add_exp}
\resizebox{\columnwidth}{!}{%
\tiny
\begin{tabular}{ccccccccccccc}
\cline{2-13}
              & \multicolumn{6}{c}{\textbf{FICO}}                           & \multicolumn{6}{c}{\textbf{NHANESI}}  \\ \cline{2-13} 
 & \multicolumn{2}{c}{Acc} & \multicolumn{2}{c}{Psb} & \multicolumn{2}{c|}{Sps} & \multicolumn{2}{c}{Acc} & \multicolumn{2}{c}{Psb} & \multicolumn{2}{c}{Sps} \\ \cline{2-13} 
              & Pos  & Neg  & Pos  & Neg  & Pos & \multicolumn{1}{c|}{Neg}  & Pos  & Neg  & Pos  & Neg  & Pos & Neg \\
\textbf{L-CR} & 0.98 & 0.94 & 0.98 & 0.99 & 5   & \multicolumn{1}{c|}{5}    & 0.99 & 0.98 & 0.98 & 0.97 & 5   & 6   \\
\textbf{R-CR} & 0.90 & 0.94 & 0.98 & 0.99 & 9   & \multicolumn{1}{c|}{8.43} & 0.86 & 0.95 & 0.96 & 0.99 & 7   & 7   \\
\textbf{AReS} & 0.34 & 0.01 & 0.85 & 0.86 & 2   & \multicolumn{1}{c|}{1}    & 0.06 & 1    & 0.87 & 0.92 & 1   & 1   \\
\textbf{CET}  & 0.76 & 0    & 0.76 & 0.60 & 2   & \multicolumn{1}{c|}{2}    & 0    & 0.40 & 0.82 & 0.56 & 0   & 5  
\end{tabular}%
}
\end{table}


\section{Parameters detailed}
In this section, we give the different parameters of each method. For all methods and datasets, we first used a greedy search given a set of parameters. For AReS, we use the following set of parameters:
\begin{itemize}
    \item max rule = $\{4, 6, 8\}$, max rule length $=\{4, 8 \}$, max change num $= \{2, 4, 6\}$,
    \item minimal support $= 0.05$, discretization bins = $\{ 10, 20\}$,
    \item $\lambda_{acc} = \lambda_{cov} = \lambda_{cst} = 1$.
\end{itemize}

Lastly, for the Counterfactual Rules, we used the following parameters:
\begin{itemize}
    \item nb estimators = $\{20, 50 \}$, max depth= $\{8, 10, 12\}$,
    \item $\pi=0.9$, $\pi_C=0.9$.
\end{itemize}
We obtained the same optimal parameters for all datasets:
\begin{itemize}
    \item AReS:  max rule $= 4$, max rule length$= 4$, max change num $= 4$, minimal support $= 0.05$, discretization bins = $10$, $\lambda_{acc} = \lambda_{cov} = \lambda_{cst} = 1$
    \item CET: max iterations $= 1000$, max leaf size $=-1$, $\lambda = 0.01, \gamma = 1 $
    \item CR: nb estimators$= 20$, max depth$=10$, $\pi=0.9$, $\pi_C=0.9$
\end{itemize}

\end{document}